\pdfoutput=1

\documentclass[11pt]{article}

\usepackage[preprint]{acl}

\usepackage{times}
\usepackage{latexsym}

\usepackage[T1]{fontenc}

\usepackage[utf8]{inputenc}

\usepackage{microtype}

\usepackage{inconsolata}

\usepackage{graphicx}
\usepackage{amsmath}

\usepackage{hyperref}
\usepackage{url}
\usepackage{tabu}
\usepackage{booktabs}
\usepackage{multirow} 
\usepackage{xcolor}
\usepackage{wrapfig} 
\usepackage{colortbl}
\usepackage[most]{tcolorbox}
\usepackage{subcaption}
\usepackage{titletoc}
\usepackage[capitalize]{cleveref} 
\usepackage{amssymb}

\usepackage{algorithm}
\usepackage{algpseudocode}

\usepackage{anyfontsize}

\definecolor{best}{RGB}{120, 220, 130}  
\definecolor{second}{RGB}{193, 255, 193}  

\definecolor{wkyellow}{RGB}{255,241,177}
\definecolor{lightblue}{HTML}{CCE5FF}
\definecolor{lightgray}{gray}{0.9}
\definecolor{goodblue}{HTML}{0071bc}

\title{Implicit Cross-Lingual Rewarding \\ for Efficient Multilingual Preference Alignment}

\author{Wen Yang\textsuperscript{1,2}, Junhong Wu\textsuperscript{1,2}, Chen Wang\textsuperscript{1,2}, Chengqing Zong\textsuperscript{1,2}, Jiajun Zhang\textsuperscript{1,2,3} \thanks{\ \ Corresponding author} \\
        \\
        \textsuperscript{1}~School of Artificial Intelligence, University of Chinese Academy of Sciences\\
        \textsuperscript{2}~Institute of Automation, Chinese Academy of Sciences \\
        \textsuperscript{3}~Wuhan AI Research \\
        \texttt{\{yangwen2023, wujunhong2021, wangchen2020\}@ia.ac.cn} \\
        \texttt{\{cqzong, jjzhang\}@nlpr.ia.ac.cn}
        }

\begin{document}
\maketitle
\begin{abstract} 
Direct Preference Optimization (DPO) has become a prominent method for aligning Large Language Models (LLMs) with human preferences. 
While DPO has enabled significant progress in aligning English LLMs, multilingual preference alignment is hampered by data scarcity.
To address this, we propose a novel approach that \textit{captures} learned preferences from well-aligned English models by implicit rewards and \textit{transfers} them to other languages through iterative training. 
Specifically, we derive an implicit reward model from the logits of an English DPO-aligned model and its corresponding reference model. This reward model is then leveraged to annotate preference relations in cross-lingual instruction-response pairs, using English instructions to evaluate multilingual responses.
The annotated data is subsequently used for multilingual DPO fine-tuning, facilitating preference knowledge transfer from English to other languages. 
Fine-tuning Llama3 for two iterations resulted in a 12.72\% average improvement in Win Rate and a 5.97\% increase in Length Control Win Rate across all training languages on the X-AlpacaEval leaderboard. 
Our findings demonstrate that leveraging existing English-aligned models can enable efficient and effective multilingual preference alignment, significantly reducing the need for extensive multilingual preference data. The code is available at \url{https://github.com/ZNLP/Implicit-Cross-Lingual-Rewarding}.

\end{abstract}

\section{Introduction}

\begin{figure}[!h]
\centering
\includegraphics[width=0.95\linewidth]{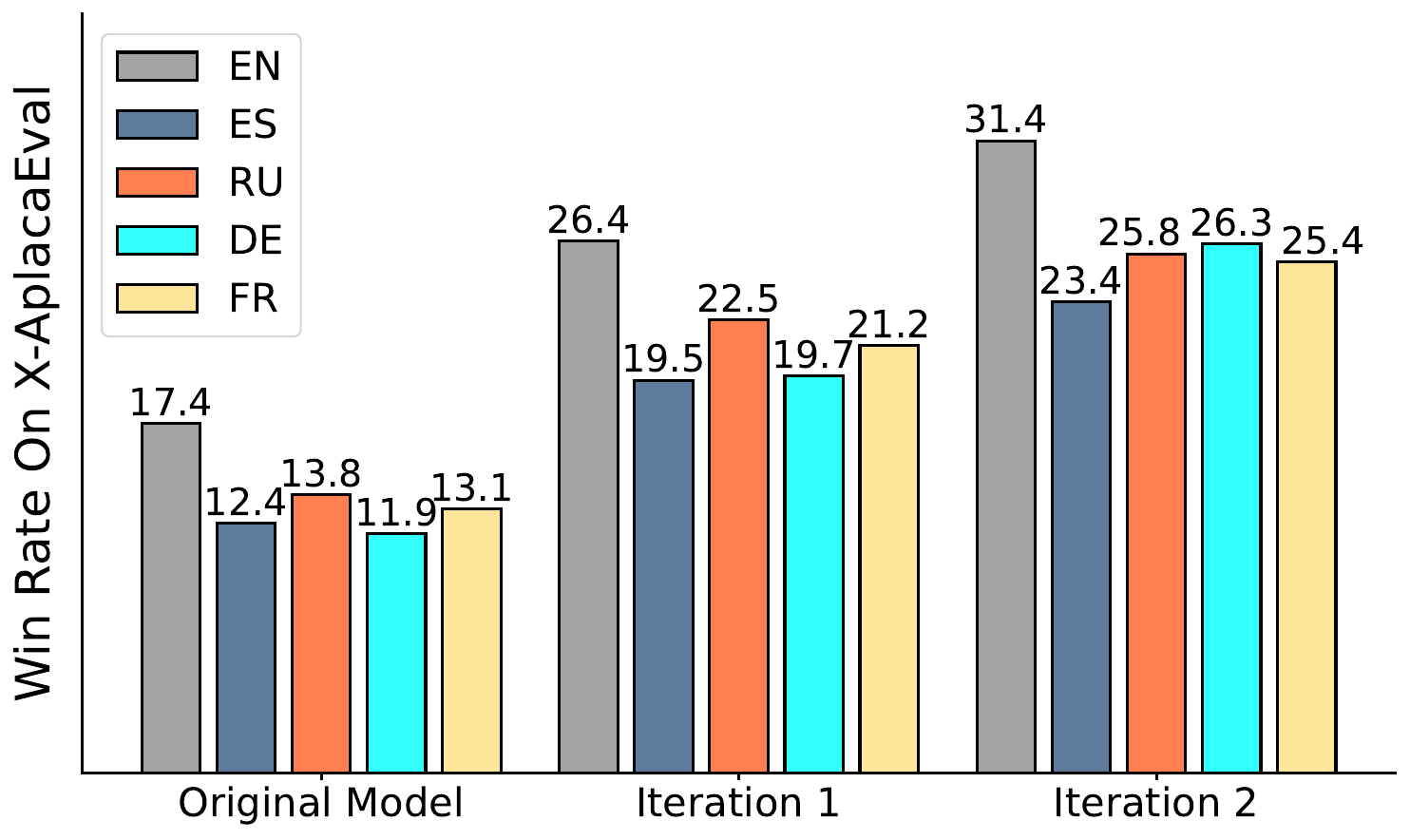}
\vspace{-2mm}
\caption{Iterative Preference Transfer and Improvement with \textbf{Implicit Cross-Lingual Rewarding} based on the English-aligned Llama3 model. Detailed results are shown in Table~\ref{tab:x_apalacaeval}.}
\vspace{-4mm}
\label{fig:comp_with_iter}
\end{figure}

\begin{figure*}[!h]
\centering
\includegraphics[width=0.95\textwidth]{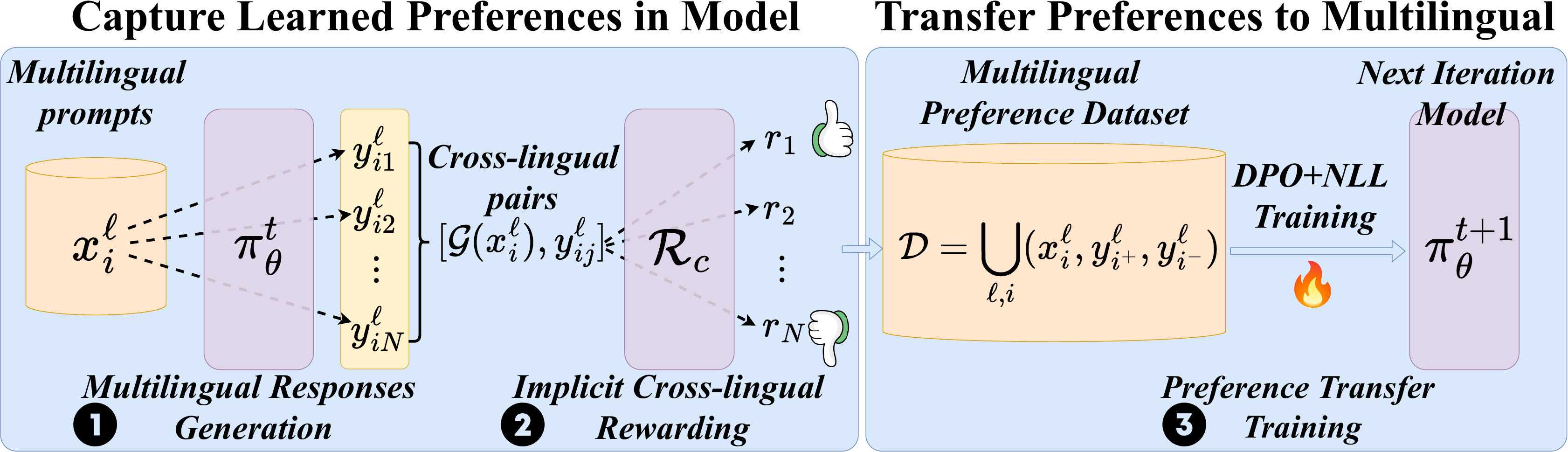}
\vspace{-2mm}
\caption{\textbf{Implicit Cross-Lingual Rewarding for Efficient Multilingual Preference Alignment.} Our method consists of three steps: (\romannumeral 1) \textit{Multilingual Responses Generation}: Sampling multilingual responses from parallel prompts with $\pi^{t}_{\theta}$, respectively. (\romannumeral 2) \textit{Implicit Cross-lingual Rewarding}: Scoring these responses with cross-lingual instruction-response pairs, where instructions are mapped into English via $\mathcal{G}(x^{\ell}_i)$ (Eq.~\ref{eq:mapping_function}) and the pairs evaluated with the implicit cross-lingual reward $\mathcal{R}_c$ (Eq.~\ref{eq:crosslingual_reward}) (\romannumeral 3) \textit{Preference Transfer Training}: Preference pairs are constructed based on scores for DPO+NLL training, producing an improved model $\pi^{t+1}_{\theta}$. This process is repeated iteratively, gradually enhancing the model’s multilingual preference alignment until optimization saturates.}
\label{fig:intro_outline}
\vspace{-4mm}
\end{figure*}

Direct alignment algorithms (DAAs), such as DPO~\citep{rafailov2024direct} and its variants~\citep{azar2024general,ethayarajh2024kto,meng2024simpo}, renowned for their simplicity, efficiency and stability than Reinforcement Learning from Human Feedback (RLHF)~\citep{ouyang2022training}, have emerged as valuable and widely adopted post-training techniques for aligning LLMs with human preferences.
While English benefits from abundant high-quality preference datasets~\citep{cui2024ultrafeedback, mukherjee2023orca} and has merged numerous DAAs-aligned models, multilingual preference alignment is challenged by data scarcity.

Existing approaches typically rely on expensive human annotation or advanced multilingual preference alignment models~\citep{ahmadian2024multilingual, dang2024rlhf} to annotate data for each language, thereby constructing off-policy multilingual preference datasets.
However, this approach faces significant challenges due to the scarcity and cost of annotations, particularly for low-resource languages. Furthermore, translation-based methods either translate English preference data into other languages~\citep{lai2023okapi} or use translation to derive reward signals to construct multilingual preference data~\citep{she2024mapo,yang2024language}. These methods can introduce artifacts and distort preference signals, hindering effective multilingual preference learning.

This work explores a novel perspective: \emph{leveraging the preference knowledge embedded within existing English-aligned models to facilitate multilingual preference alignment.}
Prior work~\citep{chen2024bootstrapping} has demonstrated that the implicit reward model, derived from the logits of a well-aligned English DPO model and its reference model, effectively captures preferences over English instructions.  
Building on this, we apply this implicit reward model to the multilingual setting, using it to label preference relations in cross-lingual instruction-response pairs~\citep{li2024x}. This ensures that multilingual responses are evaluated based on their alignment with English instructions.
We term \textit{Implicit Cross-Lingual Rewarding}, which preserves reward signal fidelity by directly evaluating multilingual responses under English instructions, avoiding translation-induced distortions.

As shown in Figure~\ref{fig:intro_outline}, our approach involves three key steps: 
(1) Multilingual response generation: Starting from any multilingual model that is DPO-tuned on English preference data from an initial reference model. Responses are sampled by the model from multilingual prompts. 
(2) Implicit cross-lingual rewarding: Constructing cross-lingual instruction-response pairs by pairing English instructions with sampled multilingual responses. The implicit reward model then assigns preference scores to these responses, capturing the model's learned preference knowledge.
(3) Preference Transfer Training: Our approach adopts iterative DPO similar to previous works~\citep{yuan2024self, yang2024language}, incorporating a negative log-likelihood (NLL) loss term to train on the multilingual preference data, thereby transferring preferences across languages.

Our experiments start with the existing English-aligned Llama 3 model, followed by two iterations of our training process. Results (Figure~\ref{fig:comp_with_iter}) demonstrate that our approach not only transfers preference knowledge from English to other languages but also iteratively improves English alignment through implicit reward. This suggests that each iteration inherently facilitates both preference transfer and refinement within the multilingual LLM.
Notably, experiments with other DAAs-aligned base models and lower-resource languages confirm the broad applicability of implicit cross-lingual rewarding, establishing it as an efficient and robust strategy for enhancing multilingual preference alignment for any English-aligned model.

\section{Preliminaries}

This section introduces two prominent methods in preference optimization, Reinforcement Learning with Human Feedback (RLHF) and Direct Preference Optimization (DPO), and derives the implicit rewards of the DPO-tuned model.

In preference optimization, the preference data typically takes the pairwise form, denoted as $\mathcal{D} = \{(x,y_w,y_l)\}$. Each prompt $ x$ is paired with two possible responses, $y_w$ and $y_l$, where $y_w$ is designated as the preferred response by human evaluators.

\subsection{Reinforcement Learning From Human Feedback}

RLHF uses human feedback to adjust a model's behavior, typically by incorporating a reward model. Since directly modeling pairwise preferences between $y_w$ and $y_l$ is difficult, a common approach defines a reward function $r(x, y)$, from which preferences are inferred, often using the Bradley-Terry model~\citep{bradley1952rank} to represent such preferences.
\begin{align}
    \label{eq:reward_objective}
    \small
    p(y_w \succ y_l|x) = \frac{\exp (r(x, y_w))}{\exp (r(x, y_w)) + \exp (r(x, y_l))}
\end{align}
From this formulation, RLHF first trains a parameterized reward model $r_{\phi}(x, y)$ using maximum likelihood:
\begin{align}
     \small
    \mathbb{E}_{(x, y_w, y_l)\sim \mathcal{D}}\left[ \log \sigma (r_{\phi}(x, y_w) - r_{\phi}(x, y_l))\right]
\end{align}
Where $\sigma$ is the logistic function, then the objective of RLHF is to optimize the policy model $\pi_{\theta}$ to maximize the expected value of the reward function. Given the pre-trained RM $r_{\phi}(x, y)$ and a reference model $\pi_{ref}$ (typically an SFT model), the objective is to find a new model $\pi_{\theta}$ by maximizing the following expression.
\begin{align}
    \label{eq:rlhf_objective}
     \small
    \max_{\pi_{\theta}} \left\{ \mathbb{E}_{\mathbf{y}\sim\pi_{\theta}(\cdot|\mathbf{x})} [ r(\mathbf{x}, \mathbf{y}) ] - \beta \log \frac{\pi_{\theta}(\mathbf{y}|\mathbf{x})}{\pi_{ref}(\mathbf{y}|\mathbf{x})} \right\}
\end{align}
Where KL divergence~\citep{kullback1951information} from the reference policy $\pi_{ref}$ is usually incorporated as a regularization to prevent the reward over-optimization of $\pi_{\theta}$, $\beta$ controls the deviation from the base reference policy. The objective is then optimized using the RL algorithm, such as Proximal Policy Optimization (PPO)~\citep{schulman2017proximal}.

\subsection{Direct Preference Optimization}

Unlike RLHF, which learns a reward model before optimizing it via reinforcement learning, Direct Preference Optimization (DPO) leverages a reward model parameterization that allows for closed-form extraction of the optimal policy, eliminating the RL training loop.  DPO's key insight is to directly model pairwise preferences.  Specifically, DPO models the probability of preferring response $y_w$ over response $y_l$ given prompt $x$ as:
\begin{equation}
\small
\begin{aligned}
\label{eq:rewrite_reward}
p_{\theta}(y_w \succ y_l | x) = \sigma \left( \beta \log \frac{\pi_{\theta}(y_w | x)}{\pi_{\text{ref}}(y_w | x)} - \beta \log \frac{\pi_{\theta}(y_l | x)}{\pi_{\text{ref}}(y_l | x)} \right)
\end{aligned}
\end{equation}
Where $\sigma$ is the sigmoid function. DPO then directly trains the optimal model on human feedback data $\mathcal{D}$ by maximizing the likelihood of these pairwise preferences using the following objective:
\begin{align}
    \label{eq:dpo_loss}
    \small
    \mathcal{L}(\pi_{\theta}) = -\mathbb{E}_{(x, y_w, y_l) \sim \mathcal{D}} \left[\log p_{\theta}(y_w \succ y_l | x) \right]
\end{align}
\paragraph{Implicit Reward in DPO-tuned Model} Thus, DPO directly implicitly learns the underlying reward function without a separate reward model training stage. The reward is parameterized in terms of the corresponding optimal policy $\pi_{\theta}$ and a reference policy $\pi_{ref}$:
\begin{align}
\label{eq:implicit_reward}
\small
r(x, y) &= \beta \log \frac{\pi_{\theta}(y \mid x)}{\pi_{\text{ref}}(y \mid x)}
\end{align}

\section{Implicit Cross-Lingual Rewarding For Efficient Multilingual Alignment}

Our approach leverages an existing English preference-aligned model and multilingual training prompts to iteratively improve preference alignment across all languages without external annotations. By exploiting the model's English preference alignment capabilities, we use implicit cross-lingual rewards to progressively enhance multilingual alignment. 
For illustrative purposes, we begin with DPO in our approach and then extend to other DAA in our experiments.
The outline is shown in Figure~\ref{fig:intro_outline}, each iteration involves (1) sampling multilingual responses, (2) scoring these responses with implicit cross-lingual rewards, and (3) constructing multilingual preference pairs for DPO training.

\paragraph{Initialization}

Given any multilingual LLM $\pi^{0}_{\theta}$, that is DPO-tuned on English preference data from an initial reference model $\pi_{ref}$, and a set of parallel multilingual instructions $\mathcal{X}$, where $\mathcal{X}$ consists of English and other language instructions ($x^{en},x^{es}, \dots, x^{ru}$). After $T$ rounds training, the model is represented as $\pi^{1}_{\theta}, \pi^{2}_{\theta}, \dots, \pi^{T}_{\theta}$.

\paragraph{Multilingual Responses Generation}

For each round $t \in \{1,2,\dots\}$, given input $x^{\ell}_i$, we sample $N$ responses, $y^{\ell}_{1 \dots N}$ from the model $\pi^{t}_{\theta}$, where $\pi^{t}_{\theta}$ is the latest policy model. Note that $\ell$ refers to any language supported by the model.
\begin{align}
    \label{eq:sample}
    y^{\ell}_{1 \dots N} \sim \pi^{t}_{\theta}(x^{\ell}_i) \quad \text{for all $x^{\ell}_i \in \mathcal{X}$}
\end{align}

\paragraph{Implicit Cross-Lingual Rewarding}
\label{sec:implict_cross_lingual_rewarding}
For any LLM $\pi_{\theta}$ that has undergone direct alignment optimization in English, the resulting model implicitly embodies a reward model. 
The implicit reward model, denoted as $r(x,y)$, can be expressed in terms of $\pi_{\theta}$ and its reference model $\pi_{ref}$, as shown in Eq.~(\ref{eq:implicit_reward}).

To leverage the learned preference in $\pi_{\theta}$, we introduce a cross-lingual reward mechanism to effectively annotate multilingual preference data using implicit rewards. For responses generated from prompts in other languages, we create cross-lingual instruction-response pairs using parallel English prompts and leverage $r(x,y)$ to score these pairs. 

Specifically, we define a mapping function $\mathcal{G}: \mathcal{I}^{\ell} \to \mathcal{I}^{en}$, where $\mathcal{I}^{\ell}$ represents the space of instructions in language $\ell$, and $\mathcal{I}^{\text{en}}$ represents the space of English instructions. Given an instruction $x_i^{\ell}$ in language $\ell$, we construct its corresponding English instruction $\mathcal{G}(x_i^{\ell})$, which is then used for reward scoring.
\begin{align}
    \label{eq:mapping_function}
    \mathcal{G}(x_i^{\ell}) = 
    \begin{cases}
    x_i^{\text{en}} & \text{if } \ell = \text{en}, \\
    \text{P}(\ell) + x_i^{\text{en}} & \text{if } \ell \neq \text{en}.
    \end{cases}
\end{align}
\begin{tcolorbox}[breakable, title={Cross-lingual Instruction Prefix $\text{P}(\ell)$}]
\label{prompt:cross_lingual_transfer}
    Please answer the following instruction using only $ \ell$ unless explicitly instructed to respond in a different language.
\end{tcolorbox}
In this formalization, when the target language $\ell$ is English ($\ell = \text{en}$), the function returns the original instruction $x_i^{en}$. When the target language $\ell$ is not English ($\ell \neq \text{en}$), the function prepends a cross-lingual instruction prefix $\text{P}(\ell)$, to the parallel English instruction $x_i^{en}$. This prefix $\text{P}(\ell)$ incorporates a language constraint, ensuring that the resulting instruction $\mathcal{G}(x_i^{\ell})$ is semantically aligned with the target language $\ell$ and compatible with the reward model.

To mitigate length exploitation~\citep{park2024disentangling}, a phenomenon observed in preference learning, we incorporate a length penalty used in RLHFlow~\citep{dong2024rlhf}. The cross-lingual reward $\mathcal{R}_{c}$ is then calculated as:
\begin{align}
    \label{eq:crosslingual_reward}
    \mathcal{R}_{c} = \beta \log \frac{\pi_{\theta}(y \mid \mathcal{G}(x^{\ell}_i))}{\pi_{ref}(y \mid \mathcal{G}(x^{\ell}_i))}-\alpha \left | y \right |
\end{align}
\paragraph{Preference Transfer Training}

For each input $x_i^\ell$ in language $\ell$ with its corresponding set of $N$ generated responses $\{y_{i1}^\ell, y_{i2}^\ell, \dots, y_{iN}^\ell\}$, we assign scores using $\mathcal{R}_c$. The responses receiving the highest and lowest scores are then selected to construct a preference tuple $(x_i^\ell, y_{i^+}^\ell, y_{i^-}^\ell)$.

The multilingual preference dataset, denoted as $\mathcal{D}$, is constructed by aggregating all preference tuples across all languages:
\begin{align}
    \label{eq:preference_dataset}
    \mathcal{D} = \bigcup_{\ell, i} (x_i^\ell, y_{i^+}^\ell, y_{i^-}^\ell)   
\end{align}
Finally, we employ a negative log-likelihood (NLL) loss term for the chosen labels in DPO loss in Eq.~(\ref{eq:dpo_loss}) to improve multilingual alignment performance. The resulting optimization objective is formulated as:
\begin{equation}
\small
\begin{aligned}
    \label{eq:dpo_loss_with_NLL}
    &\mathcal{L}^{\text{NLL}}_{\text{DPO}}(\pi_{\theta}) = - \frac{\log \pi_{\theta}(y_{i^+}^\ell|x_i^\ell)}{|y_{i^+}^\ell|} \\
    &-\log \sigma \left( \beta \log \frac{\pi_{\theta}(y_{i^+}^\ell | x_i^\ell)}{\pi_{\text{ref}}(y_{i^+}^\ell | x_i^\ell)} - \beta \log \frac{\pi_{\theta}(y_{i^-}^\ell | x_i^\ell)}{\pi_{\text{ref}}(y_{i^-}^\ell | x_i^\ell)} \right)
\end{aligned}
\end{equation}
After DPO training, the policy model $\pi_\theta^t$ is updated to $\pi_\theta$, which is then used to generate responses and score data for the subsequent iteration. The overall process of our approach is illustrated in Algorithm~\ref{alg:outline} in Appendix~\ref{appendix:algorithm_overview}.

\paragraph{Extension to Other DAA} 
KTO~\citep{ethayarajh2024kto}, inspired by prospect theory, directly optimizes generation utility, in contrast to DPO, which relies on pairwise preferences. We use an English KTO-aligned model as our base and apply KTO iteratively to explore the generalizability of our method beyond pairwise alignment.
Details of the KTO optimization process with our approach can be found in Appendix~\ref{appendix:kto}.

\section{Discussion}
In this section, we explore two key questions: \\ 
\noindent
\textbf{1.} Are there alternative forms of implicit reward?
\textbf{2.} Is cross-lingual reward effective?\\

\subsection{The Alternative Implicit Rewards}


We designed alternative implicit rewards using a DPO-tuned model under the same settings and compared the effect of different rewards in Section~\ref{sec:main_result} and Appendix~\ref{appendix:differnent_implicit_reward}.

Prior work~\citep{wu2024reuse, hong2024cross} shows that reward models trained only on English data can achieve \emph{zero-shot cross-lingual transfer}. Therefore, the most straightforward reward approach is the multilingual reward. Given prompt $x^{\ell}_i$ and corresponding response $y$, the \textbf{multilingual reward $\mathcal{R}_{m}$} is then calculated as:
\begin{align}
    \label{eq:multilingual_reward}
    \mathcal{R}_{m} = \beta \log \frac{\pi_{\theta}(y \mid x^{\ell}_i)}{\pi_{ref}(y \mid x^{\ell}_i)}-\alpha \left | y \right |
\end{align}
The alternative reward function directly leverages the English reward model by translating responses into English before applying the reward.
Given prompt $x^{\ell}_i$ and corresponding response $y$, the \textbf{Translate-to-English reward $\mathcal{R}_{t}$} is then calculated as:
\begin{align}
    \label{eq:translingual_reward}
    \mathcal{R}_{t} = \beta \log \frac{\pi_{\theta}(\mathcal{T}(\ell, y) \mid x^{en}_i)}{\pi_{ref}(\mathcal{T}(\ell, y) \mid x^{en}_i)}-\alpha \left | y \right |
\end{align}
where the mapping function $\mathcal{T}(\ell, y)$ is defined as:
\begin{align}
    \label{eq:translate_mapping_function}
    \mathcal{T}(\ell, y) = 
    \begin{cases}
    y & \text{if } \ell = \text{en}, \\
    \text{\small{LLM-Translate}}(y) & \text{if } \ell \neq \text{en}.
    \end{cases}
\end{align}
Here, $\mathcal{T}(\ell, y)$ acts as an identity function when $\ell$ is English, returning $y$. Otherwise, $\mathcal{T}(\ell, y)$ translates the response $y$ into English using the LLM's translation capabilities, the prompt is shown in~\ref{prompt:self_translate}. 
\textbf{Critically}, the $\mathcal{R}_c$ and $\mathcal{R}_t$ rewards are \textit{always} conditioned on the English instructions, either $\mathcal{G}(x^{l}_i)$ or $x^{en}_i$. This ensures that reward scoring across all languages is based on English instructions, keeping them within the reward model's effective range.

\subsection{The Effectiveness of Cross-lingual Reward}
To assess the effectiveness of the cross-lingual reward, we sampled 100 pairs per language from the preference pairs constructed by $\mathcal{R}_c$ and evaluated them using head-to-head comparisons with GPT-4o, the prompt is shown in~\ref{prompt:reward_acc}. Table~\ref{tab:reward_acc} shows the resulting reward accuracy, demonstrating a strong positive signal across all languages. 

Moreover, we assess the reward accuracy of multilingual reward $\mathcal{R}_{m}$, Translate-to-English reward $\mathcal{R}_{t}$ and natural multilingual reward model $\mathcal{R}_{n}$. We choose `allenai/tulu-v2.5-13b-chatbot-arena-2023-rm'~\citep{ivison2024unpacking} as $\mathcal{R}_{n}$, a reward model used for PPO training on the Chatbot Arena 2023 dataset~\citep{chiang2024chatbot}. 

Table~\ref{tab:reward_acc} demonstrates the zero-shot cross-lingual transfer capability of our English Implicit Reward Model, highlighting the effectiveness of Implicit Rewarding in English for multilingual instructions. We observed that evaluating translated non-English responses leads to unreliable reward scores. We hypothesize that this inaccuracy arises from assessing distorted translated data, consequently hindering performance in non-English languages. In contrast, our implicit cross-lingual reward approach exhibits a higher average reward accuracy across the evaluated languages compared to a reward model directly trained on natural multilingual preference data from Chatbot Arena.

\begin{table}[htbp]
\footnotesize
\centering
\renewcommand\arraystretch{1.2}
\setlength{\tabcolsep}{2.8mm}
\begin{tabular}{lcccccc}
    \toprule[1.2pt] 
    & \multicolumn{6}{c}{\textbf{Reward Accuracy (0-1)}} \\ 
    \cline{2-7}
    &en &es &ru &de & fr & \textbf{Avg}\\
    \midrule[0.8pt]
    \rowcolor{lightgray} 
    $\mathcal{R}_{c}$ & 0.71& 	0.61& 	0.62& 	0.67& 	0.69 & 0.66 \\
    $\mathcal{R}_{m}$ & 0.71 & 0.56 & 0.57 & 0.54 & 0.57 & 0.59 \\
    $\mathcal{R}_{t}$ & 0.71 & 0.46 & 0.47 & 0.46 & 0.48 & 0.52 \\
    $\mathcal{R}_{n}$ & 0.65 & 0.62 & 0.61 & 0.62 & 0.63 & 0.63 \\

\bottomrule[1.2pt]
\end{tabular}
\vspace{-2mm}
\caption{\label{tab:reward_acc} The reward accuracy of preference pairs.}
\vspace{-4mm}
\end{table}

\section{Experiments}
\subsection{Experimental Setup}
\paragraph{Models}

While prior work~\citep{meta2024introducing, yang2024qwen2} offers numerous English DPO-tuned instruction-following models, their RLHF training details are often closed. To ensure transparency, we use Llama-3-8B-SFT-DPO~\citep{meng2024simpo} as our initial English-aligned model. 
This model, derived from Meta-Llama-3-8B via SFT on UltraChat-200k ~\citep{ding2023enhancing} and DPO on UltraFeedback~\citep{cui2024ultrafeedback}, follows the Zephyr training pipeline~\citep{tunstall2023zephyr} using open-source data.

\paragraph{Languages}

English serves as our core training language, enabling both cross-lingual preference transfer and iterative self-improvement.  Our main experiments focus on Spanish (es), Russian (ru), German (de), and French (fr) to observe cross-lingual preference alignment. We also evaluate several low-resource languages, including Bengali (bn), Swahili (sw), and Thai (th), to assess performance in low-resource settings.

\paragraph{Datasets}
UltraFeedback \citep{cui2024ultrafeedback} is a large-scale, high-quality AI feedback dataset comprising 60K preference samples closely aligned with human preferences. We randomly sampled 3K UltraFeedback's prompts and translated them into other languages using the Google Translate API to create parallel multilingual prompts.

\paragraph{Implementation Details}

We sample $N=10$ responses per prompt using a \textit{temperature} of $0.9$ and \textit{top-p} of $1.0$ and optimized $\alpha$ to minimize the length difference between the chosen and rejected responses. See Appendix~\ref{appendix:hyperparameters} for further details.

\paragraph{Evaluation and Metrics}
We evaluated multilingual preference alignment from three aspects:\\
(1) First, we used \textbf{X-AlpacaEval Leaderboard} in~\citet{yang2024language}, a multilingual extension of AlpacaEval 2.0~\citep{alpaca_eval}, to compare the multilingual instruction-following abilities of various models. To mitigate length bias in LLM preferences, we report both standard Win Rate (WR) and length-controlled (LC) Win Rates.\\
(2) Second, we used \textbf{Multilingual MT-Bench}, a multilingual adaptation of MT-Bench \citep{zheng2024judging}, which consists of open-ended questions designed to assess conversational and instruction-following skills. GPT-4o was used to score model responses on a scale of 1 to 10. \\
(3) Finally, to assess the alignment tax, we evaluated our model on \textbf{Multilingual NLP benchmarks}, including multilingual version of MMLU \citep{hendrycks2020measuring}, HellaSwag \citep{zellers2019hellaswag}, ARC Challenge \citep{clark2018think}, and TruthfulQA \citep{lin2021truthfulqa}.

\subsection{Main Results}
\label{sec:main_result}
\paragraph{X-AlpacaEval Leaderboard}
\begin{table*}[!h]
\small
\centering
\renewcommand\arraystretch{1.2}
\setlength{\tabcolsep}{1.4mm}
\begin{tabular}{l|cc|cc|cc|cc|cc|cc}
    \toprule[1.2pt] 
    \multirow{2}{*}{\textbf{Model}} & \multicolumn{2}{c}{\textbf{en}} & \multicolumn{2}{c}{\textbf{es}} & \multicolumn{2}{c}{\textbf{ru}} & \multicolumn{2}{c}{\textbf{de}} & \multicolumn{2}{c}{\textbf{fr}} & \multicolumn{2}{c}{\textbf{Avg}} \\ 
    & \textit{LC} & \textit{WR} & \textit{LC} & \textit{WR} & \textit{LC} & \textit{WR} & \textit{LC} & \textit{WR} & \textit{LC} & \textit{WR} & \textit{LC} & \textit{WR} \\
    \midrule[0.8pt]
    \midrule[0.8pt]
    \rowcolor{lightgray} 
    \multicolumn{13}{l}{\textit{Cross-lingual Implicit Rewarding}}\\
    Llama-3-8B-SFT ($\pi_{I}$) & 9.02 & 6.25 & 6.34 & 3.77 & 3.96 & 3.28 & 3.71 & 2.62 & 4.73 & 3.26 &  5.55&	3.84\\
    Llama-3-8B-SFT-DPO ($\pi^{0}_{\theta}$) & 17.24 & 17.35 & 11.32 & 12.41 & 11.05 & 13.82 & 10.17 & 11.87 & 11.56 & 13.09 & 12.27&	13.71\\
    \hspace{18pt} Iteration 1 ($\pi^{1}_{\theta}$) & 20.46 & \cellcolor{second}26.40 & 14.52 & \cellcolor{second}19.49 & \cellcolor{second}16.00 & \cellcolor{second}22.50 & \cellcolor{second}14.54 & \cellcolor{second}19.69 & \cellcolor{second}17.08 & \cellcolor{second}21.20 & \cellcolor{second}16.52 & \cellcolor{second}21.86 \\
    \hspace{18pt} Iteration 2 ($\pi^{2}_{\theta}$) & \cellcolor{second}21.19 & \cellcolor{best}31.38 & \cellcolor{second}16.88 & \cellcolor{best}23.37 & \cellcolor{best}18.11 & \cellcolor{best}25.76 & \cellcolor{best}17.92 & \cellcolor{best}26.27 & \cellcolor{best}17.12 & \cellcolor{best}25.35 & \cellcolor{best}18.24 & \cellcolor{best}26.43\\
    Meta-Llama-3-8B-Instruct & \cellcolor{best}23.48 & 24.90 & \cellcolor{best}17.52 & 18.08 & 6.37 & 7.81 & 7.74 & 8.65 & 13.58 & 14.18 & 13.74&	14.72 \\
    \rowcolor{lightgray} 
    \multicolumn{13}{l}{\textit{Comparison: Language Imbalance Driven Rewarding}~\citep{yang2024language}}\\
    Best Model of Two Iterations & 18.69	&20.97&	13.99&	16.69&	12.68&	16.60&	11.31	&15.22	&12.86&	15.54&	13.91&	17.00\\
    \rowcolor{lightgray} 
    \multicolumn{13}{l}{\textit{Extension to Other English DAA-aligned Model}}\\
    Llama-3-8B-SFT-KTO ($\pi^{0}_{\theta}$) & 14.99 & 15.86 & 13.21 & 14.22 & 10.72 & 14.74 & 10.14 & 12.18 & 11.55 & 13.49 & 12.12 & 14.10\\
    \hspace{18pt} Iteration 1 ($\pi^{1}_{\theta}$) & 15.31 & 19.71 & 15.34 & 17.02 & 14.51 & 19.10 & 12.45 & 15.86 & 14.82 & 17.36 & 14.49 & 17.81 \\
    \hspace{18pt} Iteration 2 ($\pi^{2}_{\theta}$) & 15.19 & 21.36 & 15.39 & 16.60 & 16.13 & 19.47 & 14.47 & 17.26 & 15.25 & 17.22 & 15.29 & 18.38\\
    
    \midrule[0.6pt]
    \rowcolor{lightgray} 
    \multicolumn{13}{l}{\textit{SOTA Multilingual Models}}\\
    gpt-4o-mini & 47.33 & 45.17 & 48.56 & 44.63 & 48.53 & 47.03 & 48.54 & 44.20 & 48.03 & 44.93 & 48.20 & 45.19 \\
    gpt-4-0613 & 28.86 & 15.61 & 35.08 & 18.18 & 30.37 & 16.82 & 29.10 & 16.00 & 25.44 & 15.23 & 29.77 & 16.37 \\
    gpt-3.5-turbo-0125 & 24.50 & 11.96 & 31.79 & 14.42 & 28.21 & 13.74 & 27.82 & 12.41 & 28.71 & 12.70 & 28.21 & 13.05 \\
    Qwen2-72B-Instruct & 39.56 & 37.72 & 36.43 & 24.73 & 37.38 & 27.15 & 32.51 & 23.93 & 33.47 & 24.63 & 35.87 & 27.63 \\
    Meta-Llama-3-70B-Instruct & 36.54 & 39.74 & 30.65 & 32.58 & 7.43 & 9.14 & 8.26 & 9.48 & 23.27 & 25.20 & 21.23 & 23.23 \\
    InternLM2.5-Chat-20B & 28.08 & 31.77 & 13.98 & 16.62 & 9.42 & 11.10 & 9.08 & 11.56 & 10.98 & 13.61 & 14.31 & 16.93 \\
    Qwen2-7B-Instruct & 22.84 & 24.39 & 17.55 & 13.89 & 18.16 & 14.33 & 12.90 & 11.45 & 19.04 & 15.97 & 18.10 & 16.01 \\
    Mistral-7B-Instruct-v0.3 & 25.13 & 21.46 & 16.30 & 13.36 & 14.16 & 13.75 & 14.48 & 11.91 & 16.37 & 13.28 & 17.29 & 14.75 \\
    Aya-23-8B & 14.31 & 15.26 & 14.29 & 16.68 & 14.10 & 17.95 & 13.84 & 18.50 & 12.74 & 14.70 & 13.86 & 16.62 \\

    
    \bottomrule[1.2pt]
\end{tabular}
\caption{\label{tab:x_apalacaeval} \textbf{The X-AlpacaEval Leaderboard.} \textit{LC} and \textit{WR} denote length-controlled and standard win rate, respectively. The best and second-best scores in \textit{Cross-lingual Implicit Rewarding} are highlighted in \colorbox{best}{`Green'} and \colorbox{second}{`Lightgreen'}. The X-AlpacaEval leaderboard was introduced by~\citet{yang2024language}, which shows the win rate over GPT-4 Turbo evaluated by GPT-4.}
\vspace{-4mm}
\end{table*}

Table~\ref{tab:x_apalacaeval} shows that implicit cross-lingual rewarding enables continuous improvement in multilingual preference alignment across iterations. Average length-controlled (LC) and standard win rates (WR) increased by 5.97\% and 12.72\%, respectively. 
Furthermore, the English LC win rate steadily improves from 17.24\% to 21.19\%, confirming the effectiveness of implicit preference rewarding for bootstrapping English proficiency, as observed in~\citep{kim2024aligning, chen2024bootstrapping}. This continuous improvement in English performance strengthens the implicit cross-lingual reward, which is crucial for our method's iterative optimization. 
Remarkably, our model, trained without any manually annotated multilingual preference data, outperforms similarly sized Instruct models, including Llama-3-8B-Instruct, Qwen2-7B-Instruct, and Mistral-7B-Instruct-v0.3 (LC: 18.24\% vs. 13.74\%, 18.10\%, 17.29\%), all of which were trained with extensive annotated preference data.

\paragraph{Multilingual MT-Bench}

The MT-Bench results in Table~\ref{tab:multilingual_mt_bench} show a continual performance improvement, increasing from 6.20 for $\pi_\theta^0$ to 6.77 for $\pi_\theta^2$. This improvement stems from the strong reward signal provided by implicit cross-lingual rewarding. 
Because we use GPT-4o as the reference model, its advanced capabilities result in lower \textit{absolute} MT-Bench scores compared to GPT-4 evaluation. However, we focus on \textit{relative} score changes during iterative training.

\begin{table}[htbp]
\footnotesize
\centering
\renewcommand\arraystretch{1.2}
\setlength{\tabcolsep}{2.5mm}
\begin{tabular}{lcccccc}
    \toprule[1.2pt] 
    \multirow{2}{*}{\textbf{Model}} & \multicolumn{5}{c}{\textbf{Avg. Score (0-10)}} & \multirow{2}{*}{\textbf{Avg}} \\ 
    \cline{2-6}
    & en &es &ru &de &fr & \\
    \midrule[0.8pt]
    \rowcolor{lightgray} 
    $\pi^{0}_{\theta}$ & 6.86 & 5.96 & 6.01 & 5.93 & 6.23 & 6.20 \\
    $\pi^{1}_{\theta}$ & 6.93 & 6.61 & 6.42 & 6.76 & 6.56 & 6.66 \\
    $\pi^{2}_{\theta}$ & 7.02 & 6.96 & 6.44 & 6.75 & 6.68 & 6.77 \\
    
\bottomrule[1.2pt]
\end{tabular}
\caption{\label{tab:multilingual_mt_bench} The Multilingual MT-Bench Benchmark on Llama-3-8B-SFT-DPO, judged with GPT-4o.}
\vspace{-6mm}
\end{table}
   
\paragraph{Multilingual NLP Benchmarks}
\begin{table*}[!h]
\footnotesize
\renewcommand\arraystretch{1.2}
\setlength{\tabcolsep}{2.9mm}
\begin{center}
\begin{tabular}{lccccc}
    \toprule[1.2pt] 
    \multirow{2}{*}{\textbf{Model}} & \textit{Multilingual}  & \textit{Multilingual}  & \textit{Multilingual} & \multicolumn{2}{c}{\textit{Multilingual TruthfulQA}} \\
    \cline{5-6}
    & \textit{ARC challenge} & \textit{HellaSwag} & \textit{MMLU} & \textit{MC1} &  \textit{MC2}\\
    \midrule[0.8pt] 
    \midrule[0.8pt] 
    \rowcolor{lightgray} 
    
    Llama-3-8B-SFT & 0.4267\textsubscript{$\pm$0.0144} & 0.4974\textsubscript{$\pm$0.0051} & 0.5240\textsubscript{$\pm$0.0043} & 0.2909\textsubscript{$\pm$0.0161} & 0.4439\textsubscript{$\pm$0.0151} \\  
    Llama-3-8B-SFT-DPO ($\pi^{0}_{\theta}$) & 0.4653\textsubscript{$\pm$0.0145} & 0.5231\textsubscript{$\pm$0.0051} & 0.5349\textsubscript{$\pm$0.0043} & 0.3479\textsubscript{$\pm$0.0169} & 0.5079\textsubscript{$\pm$0.0161} \\  
    \hspace{18pt} Iteration 1 ($\pi^{1}_{\theta}$) & 0.4679\textsubscript{$\pm$0.0145} & 0.5255\textsubscript{$\pm$0.0051} & 0.5356\textsubscript{$\pm$0.0043} & 0.3489\textsubscript{$\pm$0.0169} & 0.5099\textsubscript{$\pm$0.0162} \\  
    \hspace{18pt} Iteration 2 ($\pi^{2}_{\theta}$) & 0.4674\textsubscript{$\pm$0.0145} & 0.5257\textsubscript{$\pm$0.0051} & 0.5364\textsubscript{$\pm$0.0043} & 0.3492\textsubscript{$\pm$0.0168} & 0.5092\textsubscript{$\pm$0.0163} \\  
    Meta-llama3-Instruct-8B & 0.4322\textsubscript{$\pm$0.0144} & 0.4833\textsubscript{$\pm$0.0051} & 0.5767\textsubscript{$\pm$0.0042} & 0.3403\textsubscript{$\pm$0.0168} & 0.5068\textsubscript{$\pm$0.0157} \\

\bottomrule[1.2pt]
\end{tabular}
\end{center}
\vspace{-2mm}
\caption{\label{tab:multilingual_NLP_benckmark}The Multilingual NLP Benchmarks.}
\vspace{-4mm}
\end{table*} 
To assess the potential degradation of world knowledge and commonsense reasoning during alignment, known as the ``alignment tax'', Table~\ref{tab:multilingual_NLP_benckmark} presents average results across the five training languages on four benchmarks (detailed results in Appendix~\ref{appendix:multilingual_NLP_benchmark}). The benchmark results show no performance degradation compared to the base model, indicating that our method effectively avoids introducing the alignment tax during preference optimization.

\paragraph{Comparison} 
\citet{yang2024language} proposed Language Imbalance Driven Rewarding, using language imbalance as a reward signal.  We compare this approach to the same settings on X-AlpacaEval (Table~\ref{tab:x_apalacaeval}). 
Note that we report the best model performance from two iterations, as we observed performance degradation in most languages in the second iteration of this approach.
While it improves multilingual alignment over $\pi^{0}_{\theta}$, its gains are significantly smaller than ours. We attribute this to its reliance on language imbalance and self-translation, limiting its effectiveness. Moreover, it doesn't address length bias, resulting in limited \textit{LC} gains. Further analysis is provided in the Appendix~\ref{appendix:comparison_with_LIDR}.

\paragraph{Extension to Other DAA} 
We extend our approach beyond DPO-aligned models to other Direct Alignment Algorithm (DAA), using an English KTO-aligned model as the base and applying KTO for iterative training. 
Results in Table~\ref{tab:x_apalacaeval} show our approach generalizes well to KTO-aligned models, effectively leveraging KTO for iterative optimization. A detailed analysis is provided in Appendix~\ref{appendix:kto}.

\begin{figure}[!h]
\centering
\includegraphics[width=0.95\linewidth]{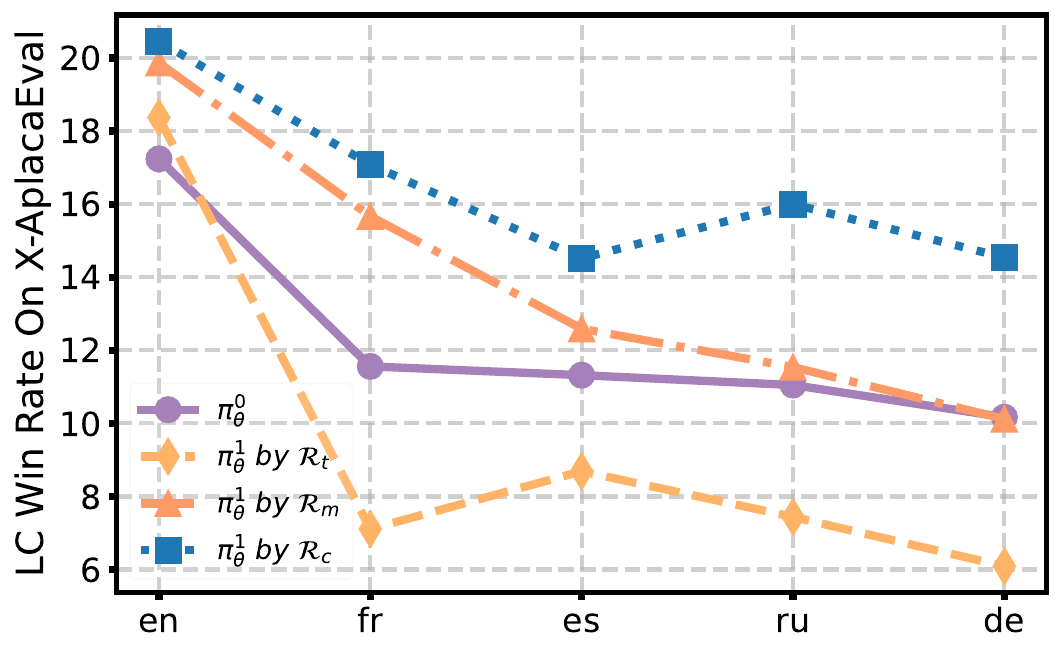}
\vspace{-2mm}
\caption{Improvement with different reward models.}
\vspace{-4mm}
\label{fig:diff_reward}
\end{figure}

\paragraph{Different Implicit Rewards}
To investigate the impact of implicit rewards on multilingual preference alignment, we compare the one-iteration performance of $\pi^{1}_{\theta}$ trained with three different implicit reward models on X-AlpacaEval (Figure~\ref{fig:diff_reward}).

The results reveal the following key findings: 
\textbf{(1)} The cross-lingual reward $\mathcal{R}_c$ yields the greatest improvement across all languages (3.20\% to 5.52\% in Table~\ref{tab:appendix_diff_rewards}). 
\textbf{(2)} The multilingual reward $\mathcal{R}_m$ demonstrates effectiveness across most languages, suggesting zero-shot cross-lingual transfer of preference alignment. However, the effectiveness of this reward is highly dependent on the model's initial proficiency in a given language (as shown in Table~\ref{tab:x_apalacaeval}). Consequently, as the model's initial proficiency decreases, the improvements conferred by the multilingual reward also diminish (as shown in Figure~\ref{fig:diff_reward}). This trend is evident in the performance gain, which ranges from 4.11\% for French to near zero for German.
\textbf{(3)} The translate-to-English reward, $\mathcal{R}_t$, degrades performance in all languages except English, suggesting that translating responses before reward evaluation is ineffective. We hypothesize that the translation process may distort the original meaning and context of the response, leading to inaccurate reward assignments and, consequently, reduced performance in non-English languages.
\textbf{(4)} While English preference data is constant, English performance is still influenced by the preference data of other languages, emphasizing the importance of high-quality preference data for each language.
Further analysis can be found in Appendix~\ref{appendix:differnent_implicit_reward}.

\subsection{More Analysis}

\paragraph{Generalization to Lower-resource Languages}

The strong performance on four high resource languages (\textit{es, ru, de, fr}) naturally raises the question: \emph{Can our method generalize to lower-resource languages?}
Experiment with Bengali (bn), Swahili (sw), Thai (th), and English (en) in Table~\ref{tab:x_apalacaeval_leaderboard_low_resource} shows the effectiveness of our approach in low-resource settings, demonstrating iterative performance gains across all languages. This is because implicit cross-lingual rewarding leverages the preference knowledge learned in English for direct (translation-free) reward, providing a strong, information-preserving reward signal for any language. 
\begin{table}[htbp]
\footnotesize
\centering
\renewcommand\arraystretch{1.2}
\setlength{\tabcolsep}{2.5mm}
\begin{tabular}{lccccc}
    \toprule[1.2pt] 
    \multirow{2}{*}{\textbf{Model}} & \multicolumn{4}{c}{\textbf{Win Rate}} & \multirow{2}{*}{\textbf{Avg}} \\ 
    \cline{2-5}
    & en &bn &sw &th & \\
    \midrule[0.8pt]
    \rowcolor{lightgray} 
    $\pi^{0}_{\theta}$ & 17.35 & 4.35 & 3.43 & 14.17 & 9.83 \\
    $\pi^{1}_{\theta}$ & 24.48 & 10.23 & 4.98 & 27.83 & 16.88 \\
    $\pi^{2}_{\theta}$ & 32.06 & 14.09 & 6.28 & 29.55 & 20.50 \\
\bottomrule[1.2pt]
\end{tabular}
\caption{\label{tab:x_apalacaeval_leaderboard_low_resource} The X-AplacaEval Leaderboard On \textbf{Llama-3-8B-SFT-DPO} in \textbf{Lower-resource languages}.}
\vspace{-4mm}
\end{table}



\paragraph{Scaling the Number of Training Prompts} 

\begin{figure}[!h]
\centering
\includegraphics[width=0.95\linewidth]{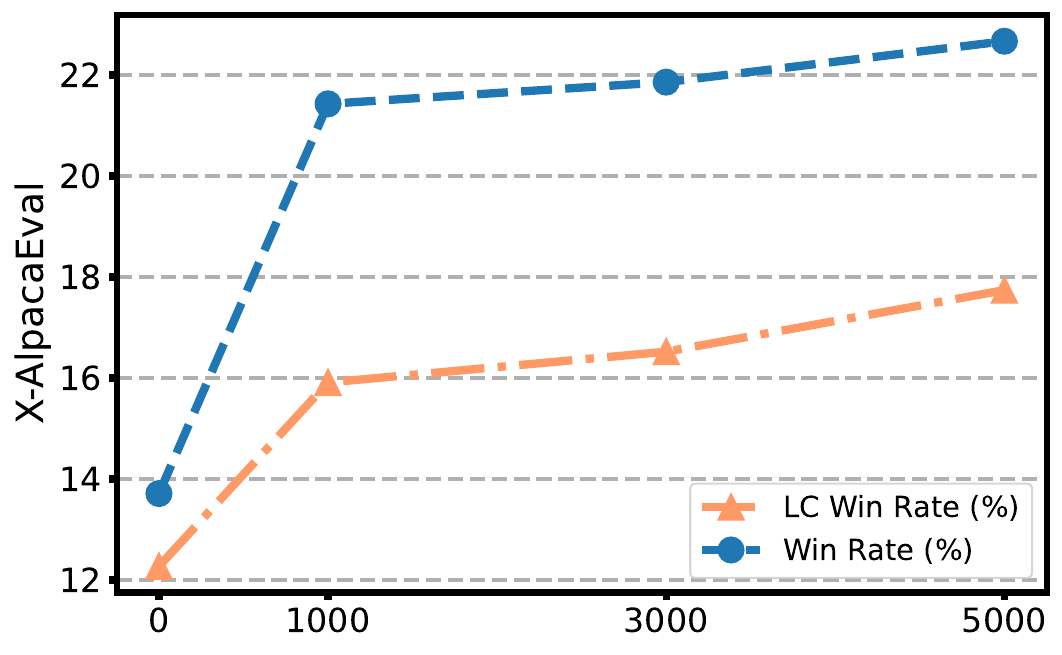}
\vspace{-2mm}
\caption{The average results for $\pi^{0}_{\theta}$ with varying training prompts.}
\vspace{-4mm}
\label{fig:prompt_num}
\end{figure}

Figure~\ref{fig:prompt_num} presents the X-AlpacaEval results for $\pi^{0}_{\theta}$ with varying training prompts in each language, demonstrating positive scaling with data volume. Notably, substantial improvements occur with as few as 1,000 prompts, a phenomenon aligned with the superficial alignment hypotheis~\citep{zhou2024lima}.
This highlights our method's efficiency and effective multilingual preference optimization with minimal data. Detailed results are provided in Appendix~\ref{appendix:scaling_the_number_of_training_prompts}.

\paragraph{Diving into Implicit Cross-Lingual Reward}
Table~\ref{tab:reward_model_ablation} investigates the impact of the cross-lingual reward defined in Eq.~(\ref{eq:crosslingual_reward}). \textbf{(1)} \textit{Effect of Length Penalty:} By adding length penalty $\alpha|y|$ in reward, the generated responses became significantly shorter (2023.8 vs. 2474.4), while the LC win rate increased by 2.27\%. While the standard win rate decreased, it is inherently susceptible to length bias. \textbf{(2)} \textit{Effect of Reference Model Selection:} 
Compared to using the previous model \(\pi^{1}_{\theta}\) as the reference, fixing the reference model to \(\pi_{I}\) improves the length-controlled (LC) win rate from 16.74\% to 18.24\% while maintaining the win rate. 
Using \(\pi_{I}\) as the reference ensures a less bias reward signal. When the previous model \(\pi^{1}_{\theta}\) is used, the reward signal can become susceptible to the evolving preferences of the model itself. This can lead to the model exploiting spurious correlations, such as length bias, rather than focusing on genuine improvements in response quality.
The less bias reward signal from \(\pi_{I}\) mitigates this issue, allowing the model to focus on generating higher-quality responses, reflected in the increased \textit{LC} win rate.
\begin{table}[htbp]
\footnotesize
\centering
\renewcommand\arraystretch{1.2}
\setlength{\tabcolsep}{2mm}
\begin{tabular}{lccc}
    \toprule[1.2pt] 
    \textbf{Different Settings} & \textit{LC} & \textit{WR} & \textit{Len} \\ 
    \midrule[0.8pt]
    \rowcolor{lightgray} 
    \multicolumn{4}{l}{\textit{Iteration 0: Initialization}}\\
    Llama-3-8B-SFT ($\pi_{I}$) & 5.55& 	3.84& 	897.6 \\
    Llama-3-8B-SFT-DPO ($\pi^{0}_{\theta}$) & 12.27&	13.71&	1695.2 \\
    \rowcolor{lightgray} 
    \multicolumn{4}{l}{\textit{Iteration 1: with / without Length Penalty $\alpha|y|$}}\\
    ($\pi^{1}_{\theta}$, Eq.~(\ref{eq:crosslingual_reward}) without $\alpha|y|$) & 14.25& 	23.07& 	2474.4 \\
    ($\pi^{1}_{\theta}$, Eq.~(\ref{eq:crosslingual_reward}) with $\alpha|y|$)  & 16.52& 	21.86& 	2023.8 \\

    \rowcolor{lightgray} 
    \multicolumn{4}{l}{\textit{Iteration 2: with Different Reference Model}}\\
    ($\pi^{2}_{\theta}$, Eq.~(\ref{eq:crosslingual_reward}) with $\pi^{1}_{\theta}$ as $\pi_{ref}$) & 16.74 & 26.62 & 2371.2 \\
    ($\pi^{2}_{\theta}$, Eq.~(\ref{eq:crosslingual_reward}) with $\pi_{I}$ as $\pi_{ref}$)  & 18.24 & 26.43 & 2254.6  \\
\bottomrule[1.2pt]
\end{tabular}
\caption{\label{tab:reward_model_ablation} The Impact of Cross-lingual Reward.}
\vspace{-6mm}
\end{table}

\section{Related Work}

\paragraph{Implicit Rewarding Optimization} 
Direct preference optimization (DPO)~\citep{rafailov2024direct} directly optimizes LLM to align with human preference by producing the optimal policy to an implicit reward model fit to the preference data.~\citet{rafailov2024r} proposed DPO within the token-level MDP setting, showing that it implicitly learns a token-level reward function using binary preference feedback.~\citet{zhong2024dpo} introduced Reinforced Token Optimization (RTO) that performs PPO based on the implicit reward in DPO.~\citet{yang2024not, chen2024cost} use implicit reward margins predicted by DPO to efficiently annotate pairwise datasets.
~\citet{chen2024bootstrapping, kim2024aligning, ko2024sera} utilized the implicit reward in the DPO-tuned model itself to construct a preference dataset and then used it in subsequent DPO rounds. 
Previous work has focused on using implicit rewards with English data in DPO-tuned models for English preference selection.  Our work introduces implicit cross-lingual rewards, leveraging English DAAs-tuned models to bootstrap capabilities across all languages.

\paragraph{Multilingual Preference Alignment}
Enhancing the multilingual capabilities of LLMs is crucial for enabling users worldwide to fully benefit from this technology~\citep{yang2023bigtranslate, li2023improving,liang2024document}.
Prior work on multilingual rewarding~\citep{wu2024reuse, hong2024cross} has explored cross-lingual transfer in reward model training using multilingual base models, showing zero-shot transfer capabilities.
Due to multilingual preference data scarcity,~\citet{ahmadian2024multilingual, dang2024rlhf} leveraged external, more powerful multilingual LLMs and reward models to construct multilingual preference data and applied optimization algorithms for multilingual alignment, incurring significant computational cost.
MAPO~\citep{she2024mapo} uses an external translation model as a reward model, aligning non-dominant languages with dominant ones by assessing consistency. However, the translator's limited context window may restrict it to other tasks.
~\citet{zhao2025mpo} introduced multilingual reward gap optimization to improve multilingual safety alignment.
~\citet{yang2024language} utilized the inherent language imbalance within LLMs to generate rewards and self-improve multilingual performance; however, this approach yields relatively coarse reward signals.
Our work addresses these limitations by using implicit cross-lingual rewarding with fine-grained reward signals to create paired data for self-iterative DPO training.

\section{Conclusion}

This paper proposes a simple yet effective framework that leverages the implicit reward model of English-aligned models as a fine-grained reward signal to bootstrap multilingual LLM alignment through a self-improving process.
Our key insight is to directly leverage English-aligned models and introduce an implicit cross-lingual reward mechanism to generate preference labels, thereby explicitly capturing preference knowledge from aligned model. 
This labeled preference data is then used to fine-tune the model itself via direct alignment algorithms, enabling the transfer and refinement of preferences from English to other languages.
Experimental results based on Llama3 demonstrate that our approach significantly improves multilingual preference alignment without any annotation data. 
This work offers a novel and efficient pathway for multilingual preference alignment.

\section*{Acknowledgements}

We thank our colleagues Chong Li, Yupu Liang, and Jianghao Chen for their valuable suggestions during the writing of this paper. We also thank the anonymous reviewers for their insightful feedback and constructive comments. 
This work is supported by National Key R\&D Program of China 2022ZD0160602 and the Strategic Priority Research Program of Chinese Academy of Sciences under Grant XDA04080400. 

\section*{Limitations}

Our work directly leverages the implicit cross-lingual reward derived from existing English-aligned models to iteratively improve the multilingual preference alignment of the model itself. 
The accuracy of the implicit cross-lingual reward significantly impacts the alignment effectiveness. If the reward signal is inaccurate or biased, it may lead to suboptimal preference optimization and hinder multilingual preference alignment. 
However, this is a common challenge in preference optimization, as RLHF also faces similar issues when the reward model is not accurate.
Another limitation is that our work focuses on general multilingual preference alignment. Developing more language-specific alignment, such as cultural alignment, is an area we plan to explore in future work.

\section*{Ethical Considerations}

This work leverages the implicit reward model of English-aligned models as a fine-grained reward signal to bootstrap multilingual LLM alignment through a self-improving process, making a novel and significant contribution to multilingual preference alignment. 
This work is dedicated to the field of efficient multilingual preference alignment, improving the alignment of large models with human preferences in multiple languages, making them better used globally.
Our contributions are entirely methodological.
Therefore, this work does not have direct negative social impacts.
In our experiments, we used publicly available datasets widely employed in prior research, containing no sensitive information to the best of our knowledge. The authors have followed ACL ethical guidelines, and the application of this work poses no apparent ethical risks.

\bibliography{custom}
\clearpage
\onecolumn
\appendix
\section*{Appendix}

\startcontents[sections]
\printcontents[sections]{l}{1}{\setcounter{tocdepth}{2}}

\clearpage
\twocolumn

\section{Implicit Cross-Lingual Rewarding}
\subsection{Base Model Setup}
\label{appendix:base_model_setup}
Prior work~\citep{meta2024introducing, yang2024qwen2} has provided numerous instruction-following models fine-tuned with DAAs on English preference data. However, the RLHF procedures for most of these models are not publicly disclosed, making it unclear whether they were trained with preference data from other languages during the DPO stage. To thoroughly explore the effectiveness of our approach, we choose Llama-3-8B-SFT-DPO\footnote{\url{https://huggingface.co/princeton-nlp/Llama-3-Base-8B-SFT-DPO}} provided by~\citet{meng2024simpo} as our initial English-aligned model. Meta-Llama-3-8B is fine-tuned on UltraChat-200k~\citep{ding2023enhancing}, resulting in Llama-3-8B-SFT\footnote{\url{https://huggingface.co/princeton-nlp/Llama-3-Base-8B-SFT}}. This model is then further optimized using Direct Preference Optimization (DPO) on UltraFeedback~\citep{cui2024ultrafeedback}, yielding the final model, Llama-3-8B-SFT-DPO. The training pipeline of Llama-3-8B-SFT-DPO follows the recipe of Zephyr~\citep{tunstall2023zephyr} and is trained on open-resource data, ensuring a high level of transparency.

Furthermore, \citet{meng2024simpo} provides models optimized with other Direct Alignment Algorithm (DAA) under the same data and training recipe. We choose Llama-3-8B-SFT-KTO\footnote{\url{https://huggingface.co/princeton-nlp/Llama-3-Base-8B-SFT-KTO}} as the base policy model to extend our approach to other English DAA-aligned models.

\subsection{Algorithm Overview}
\label{appendix:algorithm_overview}
Algorithm~\ref{alg:outline} outlines our proposed Implicit Cross-lingual Rewarding framework.  The algorithm takes as input an initial model ($\pi_I$), an English-aligned model ($\pi^0_\theta$) trained with DPO using $\pi_I$, the number of iterations ($T$), and a set of parallel multilingual prompts ($\mathcal{X}$). The core idea is to iteratively refine the multilingual preference alignment of an existing English-aligned model by leveraging its inherent English preference alignment. In each iteration $t$, preference data ($\mathcal{D}_t$) is synthesized using the implicit cross-lingual reward $\mathcal{R}_c$, derived from the previous iteration's model ($\pi^{t-1}_\theta$), the initial model ($\pi_I$). This data generation process involves calculating cross-lingual rewards (as detailed in Eq.~\ref{eq:sample}, \ref{eq:crosslingual_reward}, and \ref{eq:preference_dataset}).  Then, policy and reference models are initialized.  For each mini-batch sampled from the preference data, a training loss based on a refined DPO loss incorporating negative log-likelihood (NLL) (Eq.~\ref{eq:dpo_loss_with_NLL}) is calculated.  The model parameters are then updated using gradient descent.  After processing all mini-batches, the model for the next iteration ($\pi^t_\theta$) is initialized with the updated parameters.  This process repeats for $T$ iterations, and the final multilingual aligned model ($\pi^T_\theta$) is returned.

\begin{algorithm*}[t!]
   \caption{Implicit Cross-lingual Rewarding}
   \label{alg:outline}
\begin{algorithmic}
  \State
  \textbf{Input:} Initial model $\pi_{I}$, $\pi^{0}_{\theta}$ is English-aligned model using DPO with $\pi_{I}$,  Iterations $T$, Parallel multilingual prompts $\mathcal{X}$
  \vspace{0.05in} 
  \hrule
  \vspace{0.05in}  
  \State 
  \For{$t=1$ {\bfseries to} $T$}
  \State Sampling responses $\mathcal{Y}$ with $\pi^{t-1}_{\theta} \text{ and } \mathcal{X}$ (Eq.~\ref{eq:sample})
  \State Synthesizing preference data $\mathcal{D}_{t}$ by score $\mathcal{Y}$ with $\mathcal{R}_c$, derived from $\pi^{t-1}_{\theta}, \pi_{I}$ (Eq. \ref{eq:sample}, \ref{eq:crosslingual_reward} and \ref{eq:preference_dataset})
  \State Initialization of policy and reference models $\pi_{\theta} \leftarrow \pi^{t-1}_{\theta}$, $\pi_{ref} \leftarrow \pi^{t-1}_{\theta}$  
  \For{mini-batch $B \sim \mathcal{D}_{t}$}
  \State Calculate training loss $\mathcal{L}^{\text{NLL}}_{\text{DPO}}(\pi_{\theta})$ with refined DPO loss incorporating NLL (Eq.~\ref{eq:dpo_loss_with_NLL})
  \State Update model parameter: $\theta \leftarrow \theta - \eta  \nabla_{\theta}\mathcal{L}^{\text{NLL}}_{\text{DPO}}(\pi_{\theta})$
  \EndFor
  \State Initializing next iteration model $\pi^{t}_{\theta}$ with the updated parameters $\theta$
  \EndFor
  \State \textbf{return}~~$\pi^{T}_{\theta}$
\end{algorithmic}
\end{algorithm*}

\subsection{Optimize the Length control $\alpha$ in reward}
\label{appendix:optimize_length_control}
In our reward function $\mathcal{R}_c$, we incorporate a length penalty term, $\alpha|y|$, to discourage the generation of overly long outputs.  Subtracting this term incentivizes the model to produce concise and appropriately sized responses. The hyperparameter $\alpha$ controls the strength of this penalty; larger values of $\alpha$ impose stronger penalties for longer outputs. Following the approach in~\citet{chen2024bootstrapping},ine extend it to the multilingual setting and optimize $\alpha$ for each language $\ell$ by minimizing the expected difference in length between preferred ($y^{\ell}_{+}$) and dispreferred ($y^{\ell}_{-}$) responses within our dataset $\mathcal{D}$:
\begin{align}
    \hat{\alpha}^{\ell}=\arg \min_{a}|\mathbb{E}_{(x^{\ell},y^{\ell}_{+},y^{\ell}_{-}) \sim \mathcal{D}}(|y^{\ell}_{+}|-|y^{\ell}_{-}|)|
\end{align}
This optimization aims to find the $\alpha$ that best balances response quality and length.

\subsection{The Format of Different rewards}
\begin{figure*}[!h]
\centering
\includegraphics[width=0.95\textwidth]{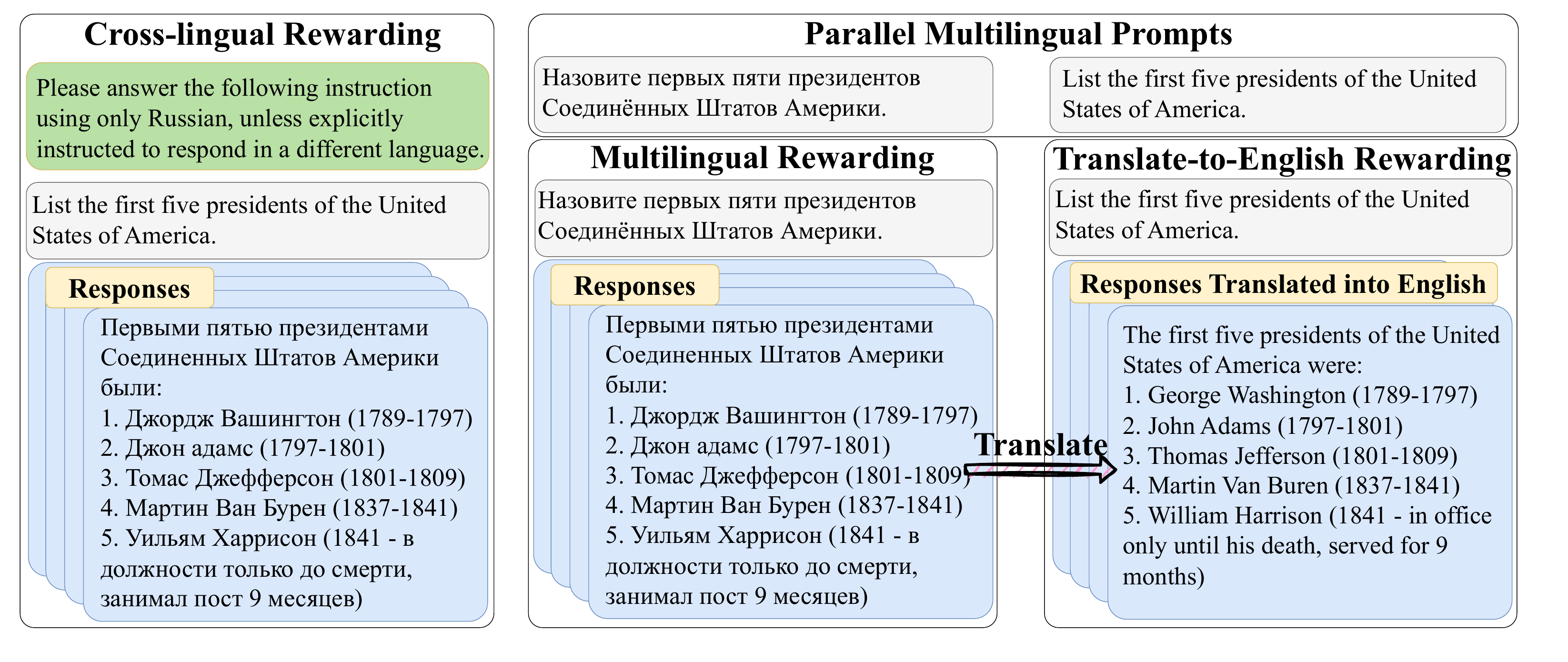}
\vspace{-2mm}
\caption{\textbf{The Format of Different Rewards.}}
\label{fig:rewarding_format}
\vspace{-4mm}
\end{figure*}

We present the data format for the cross-lingual reward, multilingual reward, and Translate-to-English reward in Figure~\ref{fig:rewarding_format}, providing a detailed breakdown of how each reward is structured and utilized within our approach to facilitate multilingual preference alignment.

\subsection{Comparison with \textit{Language Imbalance Driven Rewarding}}
\label{appendix:comparison_with_LIDR}
\citet{yang2024language} proposed \textit{Language Imbalance Driven Rewarding} for multilingual self-improving, where the inherent language imbalance between dominant and non-dominant languages within LLMs is leveraged as a reward signal. Then, using LLM itself mutually translates the dominant and non-dominant language responses to construct multilingual preference data. While the premise of language imbalance driven rewarding is compelling, and its effectiveness was demonstrated with the Llama-3-8B-Instruct model, this approach relies on the model's internal language imbalance and translation capabilities.

Table~\ref{tab:experimental_settings_between_lidr_and_icr} highlights the differences in experimental settings between Language Imbalance Driven Rewarding (LIDR) and Implicit Cross-lingual Rewarding (ICR), including variations in the base model, training dataset, and dataset size. To ensure a more equitable and direct comparison of the methods, we conducted three distinct comparative experiments between our Implicit Cross-lingual Rewarding (ICR) approach and the Language Imbalance Driven Rewarding (LIDR) method. 


\begin{table*}[htbp]
\footnotesize
\centering
\renewcommand\arraystretch{1.2}
\setlength{\tabcolsep}{2.5mm}
\begin{tabular}{llll}
    \toprule[1.2pt] 
    \textbf{Approach} & \textbf{Base Model} & \textbf{Training Dataset} & \textbf{Training Size} \\
    \midrule[0.8pt]
    \rowcolor{lightgray} 
    Language Imbalance Driven Rewarding & Llama3-8B-Instruct & Alpagasus & 1000 samples each language \\
    Implicit Cross-lingual Rewarding & Llama3-8B-SFT-DPO & UltraFeedback & 3000 samples each language \\
\bottomrule[1.2pt]
\end{tabular}
\caption{\label{tab:experimental_settings_between_lidr_and_icr} Experimental settings between Language Imbalance Driven Rewarding~\citep{yang2024language} and our approach.}
\end{table*}

\begin{table*}[!h]
\small
\centering
\renewcommand\arraystretch{1.2}
\setlength{\tabcolsep}{1mm}
\begin{tabular}{l|cc|cc|cc|cc|cc|cc}
    \toprule[1.2pt] 
    \multirow{2}{*}{\textbf{Model}} & \multicolumn{2}{c}{\textbf{en}} & \multicolumn{2}{c}{\textbf{es}} & \multicolumn{2}{c}{\textbf{ru}} & \multicolumn{2}{c}{\textbf{de}} & \multicolumn{2}{c}{\textbf{fr}} & \multicolumn{2}{c}{\textbf{Avg}} \\ 
    & \textit{LC} & \textit{WR} & \textit{LC} & \textit{WR} & \textit{LC} & \textit{WR} & \textit{LC} & \textit{WR} & \textit{LC} & \textit{WR} & \textit{LC} & \textit{WR} \\
    \midrule[0.8pt]
    \midrule[0.8pt]
    \rowcolor{lightgray} 
    \multicolumn{13}{l}{\textit{Direct comparison under their respective settings.}}\\
    Llama3-8B-Instruct (Base Model) & 23.48 & 24.90 & 17.52 & 18.08 & 6.37 & 7.81 & 7.74 & 8.65 & 13.58 & 14.18 & 13.74 & 14.72 \\
    LIDR (Iteration 1) & 26.10 & 30.11 & 18.78 & 21.82 & 14.23 & 18.01 & 14.36 & 16.87 & 14.49 & 17.51 & 17.59 & 20.86 \\
    LIDR (Iteration 2) & 27.12 & 34.09 & 15.91 & 21.21 & 13.53 & 19.25 & 12.17 & 16.02 & 15.24 & 20.34 & 16.79 & 22.18 \\
    \midrule[0.6pt]
    Llama3-8B-SFT-DPO  (Base Model) & 17.24 & 17.35 & 11.32 & 12.41 & 11.05 & 13.82 & 10.17 & 11.87 & 11.56 & 13.09 & 12.27 & 13.71 \\
    ICR (Iteraion 1) & 20.46 & 26.40 & 14.52 & 19.49 & 16.00 & 22.50 & 14.54 & 19.69 & 17.08 & 21.20 & 16.52 & 21.86 \\
    ICR (Iteraion 2) & 21.19 & 31.38 & 16.88 & 23.37 & 18.11 & 25.76 & 17.92 & 26.27 & 17.12 & 25.35 & 18.24 & 26.43 \\
    \midrule[0.8pt]
    \rowcolor{lightgray} 
    \multicolumn{13}{l}{\textit{Comparison under LIDR settings.}}\\
    Llama3-8B-Instruct (Base Model) & 23.48 & 24.90 & 17.52 & 18.08 & 6.37 & 7.81 & 7.74 & 8.65 & 13.58 & 14.18 & 13.74 & 14.72 \\
    LIDR (Iteraion 1) & 26.10 & 30.11 & 18.78 & 21.82 & 14.23 & 18.01 & 14.36 & 16.87 & 14.49 & 17.51 & 17.59 & 20.86 \\
    LIDR (Iteraion 2) & 27.12 & 34.09 & 15.91 & 21.21 & 13.53 & 19.25 & 12.17 & 16.02 & 15.24 & 20.34 & 16.79 & 22.18 \\
    \midrule[0.6pt]
    ICR (Iteraion 1) & 26.80 & 31.76 & 21.92 & 24.03 & 16.03 & 19.31 & 16.79 & 19.37 & 18.95 & 21.49 & 20.10 & 23.19 \\
    ICR (Iteraion 2) & 28.12 & 35.32 & 23.21 & 26.37 & 19.17 & 23.01 & 19.12 & 24.69 & 20.46 & 26.82 & 22.02 & 27.24 \\
    \midrule[0.8pt]
    \rowcolor{lightgray} 
    \multicolumn{13}{l}{\textit{Comparison under ICR settings.}}\\
    Llama3-8B-SFT-DPO  (Base Model) & 17.24 & 17.35 & 11.32 & 12.41 & 11.05 & 13.82 & 10.17 & 11.87 & 11.56 & 13.09 & 12.27 & 13.71 \\
    LIDR$^*$ (Best of Two Iterations) & 18.69 & 20.97 & 13.99 & 16.69 & 12.68 & 16.60 & 11.31 & 15.22 & 12.86 & 15.54 & 13.91 & 17.00 \\
    \midrule[0.6pt]
    ICR (Iteraion 1) & 20.46 & 26.40 & 14.52 & 19.49 & 16.00 & 22.50 & 14.54 & 19.69 & 17.08 & 21.20 & 16.52 & 21.86 \\
    ICR (Iteraion 2) & 21.19 & 31.38 & 16.88 & 23.37 & 18.11 & 25.76 & 17.92 & 26.27 & 17.12 & 25.35 & 18.24 & 26.43 \\
    
    \bottomrule[1.2pt]
\end{tabular}
\caption{\label{tab:x_apalacaeval_comparison_between_LIDR_and_ICR} \textbf{Comparisons between Language Imbalance Driven Rewarding (LIDR)~\citep{yang2024language} and Implicit Cross-lingual Rewarding (ICR) under various experimental settings on the X-AlpacaEval Leaderboard.} $^*$ notes that we report the best performance of the LIDR approach across two iterations under the ICR setting, as the second iteration showed performance degradation in most languages.}
\end{table*}

First, a direct comparison of the reported results within their respective experimental settings (the \texttt{upper} of Table~\ref{tab:x_apalacaeval_comparison_between_LIDR_and_ICR}) shows that our method consistently outperforms LIDR excepted for English, achieving higher average performance across five training languages after only two iterations, even when using their original base models and configurations. 

Then, we reproduced LIDR’s results under our ICR experimental setup. These reproduced results (the \texttt{middle} of Table~\ref{tab:x_apalacaeval_comparison_between_LIDR_and_ICR}) reveal performance degradation for LIDR in most languages during the second iteration, resulting in substantially smaller overall gains compared to ICR. 

We evaluated our approach under ICR’s experimental conditions, specifically employing Llama3-8B-SFT as the base model within our ICR framework. As shown at the \texttt{bottom} of Table~\ref{tab:x_apalacaeval_comparison_between_LIDR_and_ICR}, our method consistently surpasses LIDR’s performance even under their settings, demonstrating the robustness and effectiveness of the ICR approach.

\subsection{Extension to Other English DAA-aligned Model}
\label{appendix:kto}
\citet{ethayarajh2024kto} proposed KTO, inspired by Kahneman and Tversky's prospect theory \citep{tversky1992advances}, to directly maximize the utility of LLM generations rather than the log-likelihood of references. Unlike standard DPO and its variants, KTO eliminates the need for pairwise preferences, requiring only a binary signal indicating whether an output is desirable or undesirable for a given input. Therefore, we use an English KTO-aligned model as the base model and apply KTO for iterative optimization to investigate whether our method generalizes to non-pairwise direct alignment algorithms. 

The KTO training loss is provided in the following:
\begin{equation}
\small
\begin{aligned}
    \label{eq:kto_loss}
    &\mathcal{L}(\pi_{\theta}) = -\mathbb{E}_{(x, y) \sim \mathcal{D}} \left[\lambda_{y} - v(x, y) \right], \\
    &v(x, y) =  
    \begin{cases}
        \lambda_{w} \sigma(\beta \log \frac{\pi_{\theta}(y_w|x)}{\pi_{ref}(y_w|x)} - z_{ref}),\text{ if } y\sim y_{w}|x, \\
        \lambda_{l} \sigma(z_{ref} - \beta \log \frac{\pi_{\theta}(y_w|x)}{\pi_{ref}(y_w|x)} ),\text{ if } y\sim y_{l}|x.
    \end{cases} \\
    &z_{ref} = \text{KL}(\pi_{\theta}(y|x) \| \pi_{\text{ref}}(y|x)).
\end{aligned}
\end{equation}
where $\lambda_{y}$ denotes $\lambda_{w}$ for desirable response and $\lambda_{l}$ for undesirable response.

The implicit reward in KTO is derived in Eq.~(\ref{eq:kto_loss}):
\begin{align}
    r(x, y)= \log \frac{\pi_{\theta}(y_w|x)}{\pi_{ref}(y_w|x)}
\end{align}
The reward function $r(x,y)$ derived in KTO is the same as that derived in DPO. Starting from Llama3-SFT-KTO, we use Algorithm~\ref{alg:outline}, modifying the loss to the KTO loss, to perform iterative multilingual preference optimization based on the English KTO-aligned model.

The results in Table~\ref{tab:x_apalacaeval} show that after two iterations, the model's average Win Rate (WR) improved by 4.28\%, and the average Length Control (LC) win rate improved by 3.17\%. These results demonstrate the good generalization of our approach to other DAA-tuned models in English.

While multilingual preference optimization performs better starting from an English DPO-aligned base model, due to differences in optimization algorithm performance and initial policy model capabilities, the effectiveness observed with KTO demonstrates that our approach can also achieve gains with weaker-aligned models.

\section{Implementation Details}
\label{appendix:implementation_details}
\subsection{Evaluation Details}
\label{appendix:evaluation_details}

\paragraph{X-AlpacaEval Leaderboard}
\cite{zhang2023plug} introduced the X-AlpacaEval benchmark, translated into Chinese, Korean, Italian, and Spanish by professional translators.~\citet{yang2024language} extended this benchmark to include German and Russian, and introduced the X-AlpacaEval Leaderboard, thereby expanding the original English-only AlpacaEval 2.0~\citep{alpaca_eval} into a multilingual framework. We use the same prompts and configurations from X-AlpacaEval, as described in~\citet{yang2024language}, to evaluate the multilingual instruction-following capabilities of LLMs. To mitigate length bias in LLM preferences, we report both standard and length-controlled (LC) win rates. The LC win rate is calculated using a separate regression model that isolates the impact of response quality by discounting the influence of length.

\paragraph{Multilingual MT-Bench}
~\citep{zheng2024judging} includes 80 open-ended questions that evaluate a chatbot’s multi-turn conversational and instruction-following ability with human preference. 
We utilize the Multilingual MT-Bench from~\citet{yang2024language}, which collected multilingual MT-Bench datasets including German, French,  Russian, and Spanish.
Specifically, we use \texttt{GPT-4o-2024-08-06} as our judge model and to generate reference outputs due to its advanced multilingual capabilities, ensuring more accurate evaluations. Because we use GPT-4o as the reference model, its advanced capabilities result in lower \textit{absolute} MT-Bench scores compared to evaluations using GPT-4. However, our focus remains on the \textit{relative} score changes observed throughout the iterative training process.

\paragraph{Multilingual NLP Benchmark}
We used the \texttt{lm-evaluation-harness} framework~\citep{eval-harness} to evaluate changes in world knowledge, commonsense reasoning, and honesty during the multilingual preference alignment iterations. Specifically, we chose the MMLU~\citep{hendrycks2020measuring}\footnote{\url{https://huggingface.co/datasets/alexandrainst/m_mmlu}}, HellaSwag~\citep{zellers2019hellaswag}\footnote{\url{https://huggingface.co/datasets/alexandrainst/m_hellaswag}}, ARC Challenge~\citep{clark2018think}\footnote{\url{https://huggingface.co/datasets/alexandrainst/m_arc}} and TruthfulQA~\citep{lin2021truthfulqa}\footnote{\url{https://huggingface.co/datasets/alexandrainst/m_truthfulqa}} benchmarks, using the multilingual versions provided by Okapi~\citep{lai2023okapi}. These multilingual benchmarks were created by translating the original benchmarks using ChatGPT.
We list the detailed information of the benchmarks as follows:

\textbf{MMLU (Massive Multitask Language Understanding)}: This benchmark \citep{hendrycks2020measuring} comprises 57 tasks, ranging from elementary math to law and ethics, testing a model's world knowledge and problem-solving abilities across diverse domains.

\textbf{HellaSwag}: HellaSwag~\citep{zellers2019hellaswag} is a challenging commonsense NLI benchmark focused on sentence completion. It presents multiple-choice questions where the plausible continuations require human-level commonsense inference. It is designed to be difficult for models relying on superficial statistical cues.

\textbf{The AI2 Reasoning Challenge (ARC) dataset}: The ARC dataset~\citep{clark2018think} focuses on question answering that contains questions from science exams from grade 3 to grade 9. It comprises two challenge sets: the Challenge Set, which contains the more difficult questions that require reasoning, and the Easy Set, which contains simpler questions.

\textbf{TruthfulQA}: TruthfulQA~\citep{lin2021truthfulqa} evaluates a model's ability to measure whether a language model is truthful in generating answers to questions.  It assesses whether a model can respond truthfully even when presented with misleading or deceptive information.  Because evaluating truthfulness in generation tasks is difficult, the benchmark provides two multiple-choice formats, MC1 (single-true) and MC2 (multi-true), which test the model's ability to identify true statements.

\subsection{Experimental Environments}
\label{appendix:experiments_environments}
All experiments were conducted on 8 NVIDIA A800 80G GPUs. Our code primarily relies on Python 3.10 and PyTorch 2.3.0. Models were fine-tuned with LLaMA-Factory~\citep{zheng2024llamafactory} and inference was performed with \texttt{vLLM 0.6.1}~\citep{kwon2023efficient}. Training for all models was launched with the \texttt{accelerate}~\citep{accelerate} library, utilizing \texttt{DeepSpeed ZeRO Stage 2}~\citep{rajbhandari2021zero}.

\subsection{Hyperparameters}
\label{appendix:hyperparameters}
For preference pair construction, we sample $N=10$ responses per prompt using a \textit{temperature} of $0.9$ and \textit{top-p} of $1.0$. During reward scoring, we optimized $\alpha$ to minimize the length difference between the chosen and rejected labels. For preference training, models are trained for one epoch per iteration with a learning rate of $5e-7$ and a batch size of $16$. The DPO hyperparameter $\beta$ was set to a fixed value of $0.1$ for all training runs. We employed the AdamW optimizer and a cosine learning rate scheduler with a warm-up phase corresponding to $3\%$ of the total training steps. 

\begin{table}[htbp]
\centering
\footnotesize
\setlength{\tabcolsep}{1.5mm}
\renewcommand\arraystretch{1.2}
\begin{tabular}{l|ccccc}
    \toprule[1.2pt] 
    \textbf{Experiments} & LR & BS & warm-up & Epoch & $\beta$ \\ 
    \midrule[0.8pt]
    \rowcolor{lightgray}  
    \multicolumn{6}{l}{\textit{Cross-Lingual Rewarding}} \\
    DPO-Tuned & 5e-7 & 16 & 0.03 & 1 & 0.1 \\
    KTO-Tuned & 5e-7 & 32 & 0.03& 1 & 0.1 \\
    \bottomrule[1.2pt]
\end{tabular}
\caption{\label{tab:hyperparameters} The hyperparameters on various experiments. `LR' refers to the Learning Rate, and `BS' denotes the Batch Size}
\end{table}

\section{Detailed Results and Analysis Across Languages}
In this section, we provide more fine-grained results and analyses from our experiments to facilitate a clearer observation of each language's performance.

\subsection{Multilingual NLP Benchmark}
\label{appendix:multilingual_NLP_benchmark}
Table~\ref{tab:appendix_multilingual_NLP_benckmark} presents detailed results on four multilingual NLP benchmarks. These detailed results offer insights into our method's performance across various languages and tasks. The table demonstrates that our approach maintains performance comparable to the Llama-3-8B-SFT-DPO base model, effectively avoiding the ``alignment tax'' — the phenomenon where aligning a model with human preferences can negatively impact its performance on multilingual NLP tasks. This indicates that our approach successfully balances preference alignment with the preservation of general language understanding capabilities.

\subsection{Generalization to Lower-resource Languages}
\label{appendix:appendix_generalization_to_lower_resource_languages}

Table~\ref{tab:appendix_on_low_resource} presents performance results for lower-resource languages, including Bengali (bn), Swahili (sw), and Thai (th), which generally exhibit lower performance compared to middle-resource languages like Spanish, Russian, German, and French in Llama 3. While English saw a decline in length control win rate during the second iteration, possibly due to transferred length control preferences from other languages not perfectly aligned with optimal English preferences, the consistent win rate improvements across the other languages demonstrate the effectiveness of our cross-lingual implicit rewarding approach. This suggests that our method successfully transfers learned knowledge and preferences, promoting strong generalization even in lower-resource settings.

\begin{table*}[!h]
\small
\centering
\renewcommand\arraystretch{1.2}
\setlength{\tabcolsep}{2.0mm}
\begin{tabular}{l|cc|cc|cc|cc|cc}
    \toprule[1.2pt] 
    \multirow{2}{*}{\textbf{Model}} & \multicolumn{2}{c}{\textbf{en}} & \multicolumn{2}{c}{\textbf{bn}} & \multicolumn{2}{c}{\textbf{sw}} & \multicolumn{2}{c}{\textbf{th}} & \multicolumn{2}{c}{\textbf{Avg}} \\ 
    & \textit{LC} & \textit{WR} & \textit{LC} & \textit{WR} & \textit{LC} & \textit{WR} & \textit{LC} & \textit{WR} & \textit{LC} & \textit{WR} \\
    \midrule[0.8pt]
    \midrule[0.8pt]
    \rowcolor{lightgray} 
    \multicolumn{11}{l}{\textit{Cross-lingual Implicit Rewarding}}\\
    Llama-3-8B-SFT-DPO ($\pi^{0}_{\theta}$) & 17.24 & 17.35 & 2.95 & 4.35 & 2.61 & 3.43 & 10.29 & 14.17 & 8.27 & 9.83 \\
    \hspace{18pt} Iteration 1 ($\pi^{1}_{\theta}$) & 21.96 & 24.48 & 5.85 & 10.23 & 2.82 & 4.98 & 21.52 & 27.83 & 13.04 & 16.88 \\
    \hspace{18pt} Iteration 2 ($\pi^{2}_{\theta}$) & 17.68 & 32.06 & 7.85 & 14.09 & 3.21 & 6.28 & 23.19 & 29.55 & 12.98 & 20.50 \\
    
    \bottomrule[1.2pt]
\end{tabular}

\caption{\label{tab:appendix_on_low_resource} The X-ApacaEval Leaderboard on lower resource languages. \textit{LC} and \textit{WR} denote length-controlled and standard win rate, respectively.}
\end{table*}

\subsection{Diving into Implicit Cross-Lingual Reward}
\label{appendix:diving_into_implicit_cross_lingual_reward}

Table~\ref{tab:appendix_analysis_reward} analyzes the effects of \textit{Length Penalty} and \textit{Reference Model Selection}. 

To investigate the effect of \textit{Length Penalty}, we compare controlled experiments in Iteration 1. Using optimal length penalty $\alpha|y|$ in cross-lingual rewarding minimizes the length difference between chosen and rejected responses, thereby reducing length bias in the preference data as much as possible. Compared to the setting without a length penalty, applying a length penalty improves the length control win rate across all languages. However, the shorter response length in the penalty setting also results in a slight decrease in the win rate across all languages except French.

Regarding \textit{reference model selection}, using the initial model, $\pi_I$, as a fixed reference, instead of the previous iteration's model, $\pi^1_\theta$, further reduces average generation length (from 2371.2 to 2254.6) and consequently improves the length control (LC) win rate across all languages. While French (fr) and German (de) saw improvements in win rate, the other three languages experienced a slight decrease. Using the initial model ($\pi_I$) as a reference provides a less bias reward signal.  A moving reference (like $\pi^1_\theta$) can lead to the reward signal drifting towards the model's own evolving (and potentially flawed) preferences, encouraging undesirable traits like length bias.  The stability of a fixed $\pi_I$ mitigates this, promoting higher-quality responses and improving the length-controlled win rate.

\begin{table*}[!h]
\small
\centering
\renewcommand\arraystretch{1.2}
\setlength{\tabcolsep}{0.8mm}
\begin{tabular}{l|cc|cc|cc|cc|cc|ccc}
    \toprule[1.2pt] 
    \multirow{2}{*}{\textbf{Model}} & \multicolumn{2}{c}{\textbf{en}} & \multicolumn{2}{c}{\textbf{es}} & \multicolumn{2}{c}{\textbf{ru}} & \multicolumn{2}{c}{\textbf{de}} & \multicolumn{2}{c}{\textbf{fr}} & \multicolumn{3}{c}{\textbf{Avg}} \\ 
    & \textit{LC} & \textit{WR} & \textit{LC} & \textit{WR} & \textit{LC} & \textit{WR} & \textit{LC} & \textit{WR} & \textit{LC} & \textit{WR} & \textit{LC} & \textit{WR} & \textit{Len} \\
    \midrule[0.8pt]
    \midrule[0.8pt]
    Llama-3-8B-SFT-DPO ($\pi^{0}_{\theta}$) & 17.24 & 17.35 & 11.32 & 12.41 & 11.05 & 13.82 & 10.17 & 11.87 & 11.56 & 13.09 & 12.27 & 13.71 & 1695.2  \\
    Iteration 1 ($\pi^{1}_{\theta}$) & 20.46 & 26.40 & 14.52 & 19.49 & 16.00 & 22.50 & 14.54 & 19.69 & 17.08 & 21.20 & 16.52 & 21.86 & 2023.8  \\
     \rowcolor{lightgray} 
    \multicolumn{14}{l}{\textit{Iteration 1: with / without Length Penalty $\alpha|y|$}}\\
    $\pi^{1}_{\theta}$, Eq.~(\ref{eq:crosslingual_reward}) without $\alpha|y|$ & 16.78 & 27.05 & 13.98 & 21.12 & 15.04 & 24.95 & 12.76 & 21.45 & 12.70 & 20.78 & 14.25 & 23.07 & 2474.4 \\ 
    $\pi^{1}_{\theta}$, Eq.~(\ref{eq:crosslingual_reward}) with $\alpha|y|$  & 20.46 & 26.40 & 14.52 & 19.49 & 16.00 & 22.50 & 14.54 & 19.69 & 17.08 & 21.20 & 16.52 & 21.86 & 2023.8 \\  
    \rowcolor{lightgray} 
    \multicolumn{14}{l}{\textit{Iteration 2 with Different Reference Model}}\\
    $\pi^{2}_{\theta}$, Eq.~(\ref{eq:crosslingual_reward}) with $\pi^{1}_{\theta}$ as $\pi_{ref}$ & 18.19 & 32.79 & 16.64 & 25.03 & 16.76 & 26.21 & 15.20 & 24.50 & 16.92 & 24.58 & 16.74 & 26.62 & 2371.2 \\
    $\pi^{2}_{\theta}$, Eq.~(\ref{eq:crosslingual_reward}) with $\pi_{I}$ as $\pi_{ref}$ & 21.19 & 31.38 & 16.88 & 23.37 & 18.11 & 25.76 & 17.92 & 26.27 & 17.12 & 25.35 & 18.24 & 26.43 & 2254.6  \\
    \bottomrule[1.2pt]
\end{tabular}

\caption{\label{tab:appendix_analysis_reward} The X-AlpacaEval Leaderboard on the Analysis of Cross-Lingual Reward. \textbf{\textit{Len} denotes the average character length of responses.}}
\end{table*}

\subsection{Different Implicit Rewards}
\label{appendix:differnent_implicit_reward}

\begin{table*}[!h]
\small
\centering
\renewcommand\arraystretch{1.2}
\setlength{\tabcolsep}{0.6mm}
\begin{tabular}{l|cc|cc|cc|cc|cc|ccc}
    \toprule[1.2pt] 
    \multirow{2}{*}{\textbf{Model}} & \multicolumn{2}{c}{\textbf{en}} & \multicolumn{2}{c}{\textbf{es}} & \multicolumn{2}{c}{\textbf{ru}} & \multicolumn{2}{c}{\textbf{de}} & \multicolumn{2}{c}{\textbf{fr}} & \multicolumn{3}{c}{\textbf{Avg}} \\ 
    & \textit{LC} & \textit{WR} & \textit{LC} & \textit{WR} & \textit{LC} & \textit{WR} & \textit{LC} & \textit{WR} & \textit{LC} & \textit{WR} & \textit{LC} & \textit{WR} & \textit{Len} \\
    \midrule[0.8pt]
    \midrule[0.8pt]
    Llama-3-8B-SFT-DPO ($\pi^{0}_{\theta}$) & 17.24 & 17.35 & 11.32 & 12.41 & 11.05 & 13.82 & 10.17 & 11.87 & 11.56 & 13.09 & 13.23 & 13.37 & 1695.2 \\
    \rowcolor{lightgray} 
    \multicolumn{14}{l}{\textit{Iteration 1 with Different Reward Modeling}}\\
    Translate-to-English $\mathcal{R}_t$, Eq.~(\ref{eq:translingual_reward}) & 18.37 & 24.95 & 8.69 & 12.56 & 7.44 & 12.01 & 6.09 & 10.33 & 7.11 & 12.63 & 12.41 & 13.78 & 2209.6\\
    Multilingual $\mathcal{R}_m$, Eq.~(\ref{eq:multilingual_reward}) & 19.87 & 26.86 & 12.59 & 17.79 & 11.54 & 18.86 & 10.12 & 17.33 & 15.67 & 20.58 & 17.09 & 18.21 & 2119.6 \\
    Cross-lingual $\mathcal{R}_c$, Eq.~(\ref{eq:crosslingual_reward}) & 20.46 & 26.40 & 14.52 & 19.49 & 16.00 & 22.50 & 14.54 & 19.69 & 17.08 & 21.20 & 18.92 & 20.25 & 2023.8\\
    \bottomrule[1.2pt]
\end{tabular}

\caption{\label{tab:appendix_diff_rewards} The X-AlpacaEval Leaderboard on different Implicit Rewards.}
\end{table*}

To investigate the impact of different reward modeling on multilingual preference alignment, we compare the performance of $\pi^{1}_{\theta}$ trained for one iteration using three different types of implicit reward models on X-AlpacaEval. Table~\ref{tab:appendix_diff_rewards} presents the performance of $\pi^{1}_{\theta}$ on X-AlpacaEval under three reward modeling approaches: cross-lingual reward ($\mathcal{R}_c$), multilingual rewarding ($\mathcal{R}_m$), and Translate-to-English reward ($\mathcal{R}_t$). 

The results reveal the following key findings: 

(1) The cross-lingual reward $\mathcal{R}_c$ yields the greatest improvement in preference alignment across all languages, outperforming the other reward models. Furthermore, by leveraging the initial model's English-language reward capabilities, $\mathcal{R}_c$ confers substantial gains to $\pi_\theta^1$ across all languages, ranging from 3.20\% to 5.52\% shown in Table~\ref{tab:appendix_diff_rewards}. 

(2) The multilingual reward $\mathcal{R}_m$ demonstrates effectiveness across most languages, suggesting that preference alignment learned in English can be effectively transferred to other languages in a zero-shot manner, consistent with the findings of~\citep{wu2024reuse, hong2024cross}. However, the effectiveness of the multilingual reward is highly dependent on the model's initial proficiency in a given language. As the model's initial proficiency decreases, the improvements conferred by the multilingual reward also diminish. As shown in Figure~\ref{fig:diff_reward} and Table~\ref{tab:appendix_diff_rewards}, the improvement in $\pi_{\theta}^1$ conferred by $\mathcal{R}_m$ decreases as the initial model $\pi_{\theta}^0$'s alignment capability diminishes across languages, from 4.11\% for French to near zero for German. 

(3) The Translate-to-English reward, $\mathcal{R}_t$, leads to a performance decline in all languages except English, suggesting that translating responses into English before reward evaluation is ineffective. We hypothesize that this is because the implicit reward, derived from generation probabilities, is computed on parallel English data after translation. This translation process may distort the original meaning and context of the response, leading to inaccurate reward assignments and, consequently, reduced performance in non-English languages.

(4) While the English preference data remains consistent regardless of the reward model, performance differences arise during multilingual preference optimization. Although bootstrapping English preferences with implicit rewards is effective, as shown in prior work, our findings reveal that English performance is still influenced by preference data from other languages.  Specifically, $\mathcal{R}_c$ achieves the best results, highlighting the importance of preference data quality across all languages when training multilingual models.

\subsection{Scaling the Number of Training Prompts}
\label{appendix:scaling_the_number_of_training_prompts}

Table \ref{tab:appendix_training_prompts_num} shows the effect of training set size on multilingual preference alignment performance. We can observe two points: (1) Increasing the training set size generally improved performance across most languages, although French (fr) showed signs of over-optimization when the number of training prompts rose from 3000 to 5000. (2) As the number of samples increases, the gain from the improvement becomes smaller. Using only 1000 prompts can improve LC and WR by 3.64\% and 7.72\%, respectively, while from 1000 to 5000, it only improves LC and WR by 1.83\% and 1.24\%. Our approach demonstrates efficient multilingual preference alignment, achieving strong performance with fewer training samples.

\begin{table*}[!h]
\small
\centering
\renewcommand\arraystretch{1.2}
\setlength{\tabcolsep}{0.6mm}
\begin{tabular}{l|cc|cc|cc|cc|cc|ccc}
    \toprule[1.2pt] 
    \multirow{2}{*}{\textbf{Model}} & \multicolumn{2}{c}{\textbf{en}} & \multicolumn{2}{c}{\textbf{es}} & \multicolumn{2}{c}{\textbf{ru}} & \multicolumn{2}{c}{\textbf{de}} & \multicolumn{2}{c}{\textbf{fr}} & \multicolumn{3}{c}{\textbf{Avg}} \\ 
    & \textit{LC} & \textit{WR} & \textit{LC} & \textit{WR} & \textit{LC} & \textit{WR} & \textit{LC} & \textit{WR} & \textit{LC} & \textit{WR} & \textit{LC} & \textit{WR} & \textit{Len} \\
    \midrule[0.8pt]
    \midrule[0.8pt]
    Llama-3-8B-SFT-DPO ($\pi^{0}_{\theta}$) & 17.24 & 17.35 & 11.32 & 12.41 & 11.05 & 13.82 & 10.17 & 11.87 & 11.56 & 13.09 & 13.23 & 13.37 & 1695.2 \\
    \rowcolor{lightgray} 
    \multicolumn{14}{l}{\textit{Iteration 1 with Different Training Prompts in Each language}}\\
   Llama-3-8B-SFT-DPO ($\pi^{0}_{\theta}$) & 17.24 & 17.35 & 11.32 & 12.41 & 11.05 & 13.82 & 10.17 & 11.87 & 11.56 & 13.09 & 12.27 & 13.71 & 1695.2 \\  
    $\pi^{1}_{\theta}$ with 1000 prompts & 19.93 & 23.74 & 15.27 & 20.71 & 14.19 & 20.82 & 13.84 & 19.91 & 16.34 & 21.96 & 15.91 & 21.43 & 2085.2 \\  
    $\pi^{1}_{\theta}$ with 3000 prompts & 20.46 & 26.40 & 14.52 & 19.49 & 16.00 & 22.50 & 14.54 & 19.69 & 17.08 & 21.20 & 16.52 & 21.86 & 2023.8 \\ 
    $\pi^{1}_{\theta}$ with 5000 prompts & 24.19 & 26.96 & 16.71 & 21.09 & 16.24 & 22.49 & 16.55 & 21.75 & 15.01 & 21.05 & 17.74 & 22.67 & 2099.8 \\  

    \bottomrule[1.2pt]
\end{tabular}

\caption{\label{tab:appendix_training_prompts_num} The X-AlpacaEval results on Scaling the Number of Training Prompts.}
\end{table*}

\section{Dataset License}

All models and data in our work are open-sourced. We utilize prompts from the UltraFeedback~\citep{cui2024ultrafeedback} dataset for efficient multilingual alignment. We adhere to the corresponding guidelines within the data.

\begin{table*}[!h]
\centering
\footnotesize
\renewcommand\arraystretch{1.2}
\setlength{\tabcolsep}{0.8mm}
\begin{tabular}{lcccccc}
    \toprule[1.2pt] 
    \multirow{2}{*}{\textbf{Model}} & \multicolumn{5}{c}{\textbf{Training Languages}} & \multirow{2}{*}{\textbf{Avg}} \\ 
    \cline{2-6}
    & en &es &ru &de &fr\\
    \midrule[0.8pt]
    \midrule[0.8pt]
    \rowcolor{lightgray} 
    \multicolumn{7}{c}{\textit{Multilingual ARC challenge, 0-shot}} \\
    Llama-3-8B-SFT & 0.5282\textsubscript{$\pm$0.0146} & 0.4239\textsubscript{$\pm$0.0145} & 0.3661\textsubscript{$\pm$0.0141} & 0.3772\textsubscript{$\pm$0.0142} & 0.4380\textsubscript{$\pm$0.0145} & 0.4267\textsubscript{$\pm$0.0144} \\  
    Llama-3-8B-SFT-DPO ($\pi^{0}_{\theta}$) &0.5819\textsubscript{$\pm$0.0144} & 0.4598\textsubscript{$\pm$0.0146} & 0.3995\textsubscript{$\pm$0.0143} & 0.4140\textsubscript{$\pm$0.0144} & 0.4713\textsubscript{$\pm$0.0146} & 0.4653\textsubscript{$\pm$0.0145} \\  
    \hspace{18pt} Iteration 1 ($\pi^{1}_{\theta}$) & 0.5742\textsubscript{$\pm$0.0144} & 0.4684\textsubscript{$\pm$0.0146} & 0.4021\textsubscript{$\pm$0.0143} & 0.4234\textsubscript{$\pm$0.0145} & 0.4713\textsubscript{$\pm$0.0146} & 0.4679\textsubscript{$\pm$0.0145} \\  
    \hspace{18pt} Iteration 2 ($\pi^{2}_{\theta}$) &0.5785\textsubscript{$\pm$0.0144} & 0.4624\textsubscript{$\pm$0.0146} & 0.4089\textsubscript{$\pm$0.0144} & 0.4183\textsubscript{$\pm$0.0144} & 0.4688\textsubscript{$\pm$0.0146} & 0.4674\textsubscript{$\pm$0.0145} \\  
    Meta-llama3-Instruct-8B & 0.5316\textsubscript{$\pm$0.0146} & 0.4162\textsubscript{$\pm$0.0144} & 0.3781\textsubscript{$\pm$0.0142} & 0.3978\textsubscript{$\pm$0.0143} & 0.4371\textsubscript{$\pm$0.0145} & 0.4322\textsubscript{$\pm$0.0144} \\

    \rowcolor{lightgray} 
    \multicolumn{7}{c}{\textit{Multilingual HellaSwag, 0-shot}} \\
     
    Llama-3-8B-SFT &0.6008\textsubscript{$\pm$0.0049} & 0.4997\textsubscript{$\pm$0.0052} & 0.4412\textsubscript{$\pm$0.0052} & 0.4600\textsubscript{$\pm$0.0051} & 0.4855\textsubscript{$\pm$0.0052} & 0.4974\textsubscript{$\pm$0.0051} \\  
    Llama-3-8B-SFT-DPO ($\pi^{0}_{\theta}$) &0.6292\textsubscript{$\pm$0.0048} & 0.5270\textsubscript{$\pm$0.0052} & 0.4624\textsubscript{$\pm$0.0052} & 0.4864\textsubscript{$\pm$0.0052} & 0.5104\textsubscript{$\pm$0.0052} & 0.5231\textsubscript{$\pm$0.0051} \\  
    \hspace{18pt} Iteration 1 ($\pi^{1}_{\theta}$) &0.6301\textsubscript{$\pm$0.0048} & 0.5304\textsubscript{$\pm$0.0052} & 0.4655\textsubscript{$\pm$0.0052} & 0.4899\textsubscript{$\pm$0.0052} & 0.5114\textsubscript{$\pm$0.0052} & 0.5255\textsubscript{$\pm$0.0051} \\  
    \hspace{18pt} Iteration 2 ($\pi^{2}_{\theta}$) &0.6295\textsubscript{$\pm$0.0048} & 0.5306\textsubscript{$\pm$0.0052} & 0.4655\textsubscript{$\pm$0.0052} & 0.4922\textsubscript{$\pm$0.0052} & 0.5105\textsubscript{$\pm$0.0052} & 0.5257\textsubscript{$\pm$0.0051} \\  
    Meta-llama3-Instruct-8B & 0.5764\textsubscript{$\pm$0.0049} & 0.4877\textsubscript{$\pm$0.0052} & 0.4326\textsubscript{$\pm$0.0051} & 0.4483\textsubscript{$\pm$0.0051} & 0.4715\textsubscript{$\pm$0.0052} & 0.4833\textsubscript{$\pm$0.0051} \\

    \rowcolor{lightgray} 
    \multicolumn{7}{c}{\textit{Multilingual MMLU, 5-shot}} \\
    Llama-3-8B-SFT & 0.6052\textsubscript{$\pm$0.0039} & 0.5231\textsubscript{$\pm$0.0043} & 0.4817\textsubscript{$\pm$0.0044} & 0.4997\textsubscript{$\pm$0.0043} & 0.5104\textsubscript{$\pm$0.0044} & 0.5240\textsubscript{$\pm$0.0043} \\  
    Llama-3-8B-SFT-DPO ($\pi^{0}_{\theta}$) & 0.6232\textsubscript{$\pm$0.0039} & 0.5301\textsubscript{$\pm$0.0043} & 0.4883\textsubscript{$\pm$0.0044} & 0.5108\textsubscript{$\pm$0.0043} & 0.5223\textsubscript{$\pm$0.0044} & 0.5349\textsubscript{$\pm$0.0043} \\  
    \hspace{18pt} Iteration 1 ($\pi^{1}_{\theta}$) & 0.6236\textsubscript{$\pm$0.0039} & 0.5293\textsubscript{$\pm$0.0043} & 0.4853\textsubscript{$\pm$0.0044} & 0.5103\textsubscript{$\pm$0.0043} & 0.5297\textsubscript{$\pm$0.0044} & 0.5356\textsubscript{$\pm$0.0043} \\  
    \hspace{18pt} Iteration 2 ($\pi^{2}_{\theta}$) & 0.6295\textsubscript{$\pm$0.0039} & 0.5285\textsubscript{$\pm$0.0043} & 0.4843\textsubscript{$\pm$0.0044} & 0.5108\textsubscript{$\pm$0.0043} & 0.5291\textsubscript{$\pm$0.0044} & 0.5364\textsubscript{$\pm$0.0043} \\  
    Meta-llama3-Instruct-8B & 0.6567\textsubscript{$\pm$0.0038} & 0.5771\textsubscript{$\pm$0.0043} & 0.5335\textsubscript{$\pm$0.0044} & 0.5506\textsubscript{$\pm$0.0043} & 0.5654\textsubscript{$\pm$0.0043} & 0.5767\textsubscript{$\pm$0.0042} \\

    \rowcolor{lightgray} 
    \multicolumn{7}{c}{\textit{Multilingual TruthfulQA MC1, 0-shot}} \\
    Llama-3-8B-SFT & 0.3060\textsubscript{$\pm$0.0161} & 0.2725\textsubscript{$\pm$0.0159} & 0.2919\textsubscript{$\pm$0.0162} & 0.2779\textsubscript{$\pm$0.0160} & 0.3062\textsubscript{$\pm$0.0164} & 0.2909\textsubscript{$\pm$0.0161} \\  
    Llama-3-8B-SFT-DPO ($\pi^{0}_{\theta}$) & 0.3856\textsubscript{$\pm$0.0170} & 0.3232\textsubscript{$\pm$0.0167} & 0.3452\textsubscript{$\pm$0.0169} & 0.3363\textsubscript{$\pm$0.0168} & 0.3494\textsubscript{$\pm$0.0170} & 0.3479\textsubscript{$\pm$0.0169} \\  
    \hspace{18pt} Iteration 1 ($\pi^{1}_{\theta}$) & 0.3966\textsubscript{$\pm$0.0171} & 0.3321\textsubscript{$\pm$0.0168} & 0.3363\textsubscript{$\pm$0.0168} & 0.3350\textsubscript{$\pm$0.0168} & 0.3443\textsubscript{$\pm$0.0169} & 0.3489\textsubscript{$\pm$0.0169} \\  
    \hspace{18pt} Iteration 2 ($\pi^{2}_{\theta}$) & 0.3896\textsubscript{$\pm$0.0170} & 0.3370\textsubscript{$\pm$0.0167} & 0.3378\textsubscript{$\pm$0.0168} & 0.3385\textsubscript{$\pm$0.0166} & 0.3433\textsubscript{$\pm$0.0169} & 0.3492\textsubscript{$\pm$0.0168} \\  
    Meta-llama3-Instruct-8B & 0.3611\textsubscript{$\pm$0.0168} & 0.3333\textsubscript{$\pm$0.0168} & 0.3541\textsubscript{$\pm$0.0170} & 0.3173\textsubscript{$\pm$0.0166} & 0.3355\textsubscript{$\pm$0.0168} & 0.3403\textsubscript{$\pm$0.0168} \\

    \rowcolor{lightgray} 
    \multicolumn{7}{c}{\textit{Multilingual TruthfulQA MC2, 0-shot}} \\
    Llama-3-8B-SFT & 0.4531\textsubscript{$\pm$0.0147} & 0.4194\textsubscript{$\pm$0.0150} & 0.4658\textsubscript{$\pm$0.0157} & 0.4284\textsubscript{$\pm$0.0150} & 0.4528\textsubscript{$\pm$0.0152} & 0.4439\textsubscript{$\pm$0.0151} \\  
    Llama-3-8B-SFT-DPO ($\pi^{0}_{\theta}$) & 0.5354\textsubscript{$\pm$0.0158} & 0.4811\textsubscript{$\pm$0.0162} & 0.5173\textsubscript{$\pm$0.0164} & 0.4913\textsubscript{$\pm$0.0160} & 0.5146\textsubscript{$\pm$0.0162} & 0.5079\textsubscript{$\pm$0.0161} \\  
    \hspace{18pt} Iteration 1 ($\pi^{1}_{\theta}$) & 0.5460\textsubscript{$\pm$0.0158} & 0.4848\textsubscript{$\pm$0.0163} & 0.5163\textsubscript{$\pm$0.0165} & 0.4931\textsubscript{$\pm$0.0162} & 0.5094\textsubscript{$\pm$0.0163} & 0.5099\textsubscript{$\pm$0.0162} \\  
    \hspace{18pt} Iteration 2 ($\pi^{2}_{\theta}$) & 0.5443\textsubscript{$\pm$0.0159} & 0.4773\textsubscript{$\pm$0.0164} & 0.5187\textsubscript{$\pm$0.0166} & 0.4955\textsubscript{$\pm$0.0163} & 0.5102\textsubscript{$\pm$0.0164} & 0.5092\textsubscript{$\pm$0.0163} \\ 
    Meta-llama3-Instruct-8B & 0.5171\textsubscript{$\pm$0.0152} & 0.4989\textsubscript{$\pm$0.0157} & 0.5256\textsubscript{$\pm$0.0162} & 0.4890\textsubscript{$\pm$0.0157} & 0.5033\textsubscript{$\pm$0.0158} & 0.5068\textsubscript{$\pm$0.0157} \\

\bottomrule[1.2pt]
\end{tabular}
\caption{\label{tab:appendix_multilingual_NLP_benckmark} The Detailed Results of Multilingual NLP Benchmarks.}

\end{table*}
\begin{table*}[htbp]
\footnotesize
\centering
\renewcommand\arraystretch{1.2}
\setlength{\tabcolsep}{2.5mm}
\begin{tabular}{lcccccc}
    \toprule[1.2pt] 
    \multirow{2}{*}{\textbf{Model}} & \multicolumn{5}{c}{\textbf{Avg. Score (0-10)}} & \multirow{2}{*}{\textbf{Avg}} \\ 
    \cline{2-6}
    & en &es &ru &de &fr & \\
    \midrule[0.8pt]
    \rowcolor{lightgray} 
    Llama-3-8B-SFT-DPO ($\pi^{0}_{\theta}$) & 6.86 & 5.96 & 6.01 & 5.93 & 6.23 & 6.20 \\
    \hspace{18pt} Iteration 1 ($\pi^{1}_{\theta}$) & 6.93 & 6.61 & 6.42 & 6.76 & 6.56 & 6.66 \\
     \hspace{18pt} Iteration 2 ($\pi^{2}_{\theta}$)& 7.02 & 6.96 & 6.44 & 6.75 & 6.68 & 6.77 \\
    
\bottomrule[1.2pt]
\end{tabular}
\caption{\label{tab:multilingual_mt_bench} The Multilingual MT-Bench Benchmark on Llama-3-8B-SFT-DPO, judged with GPT-4o.}
\vspace{-4mm}
\end{table*}

\clearpage
\onecolumn
\section{Prompt Template}
\subsection{Cross-lingual Instruction Prefix $\text{P}(\ell)$ in mapping function $\mathcal{G}(x_i^{\ell})$}
\begin{tcolorbox}[breakable, title={Cross-lingual Instruction Prefix $\text{P}(\ell)$}]
\label{prompt:cross_lingual_transfer_v2}
    Please answer the following instruction using only $ \ell$, unless explicitly instructed to respond in a different language.\\
\end{tcolorbox}

\subsection{LLM-Translate($y$) in mapping function $\mathcal{T}(\ell,y)$}
\begin{tcolorbox}[breakable, title={Prompt in $\text{LLM-Translate}(y)$}]
\label{prompt:self_translate}
Please translate the following sentences into \textit{English}. The input sentences are wrapped by \texttt{<sentence>} and \texttt{</sentence>}:\\
\\
\texttt{<sentence>}\\
\texttt{y} \textit{(Response to $x^{\ell}_i$)}\\
\texttt{</sentence>}
\end{tcolorbox}

\subsection{Reward Accuracy Judgement Prompt}
\begin{tcolorbox}[title={Prompt for Judging Reward Accuracy}, width=\textwidth, breakable]
\label{prompt:reward_acc}
You are a helpful following assistant whose goal is to select the preferred (least wrong) output for a given instruction in \texttt{[LANGUAGE]}.\\  
\\
Which of the following answers is the best one for given instruction in \texttt{[LANGUAGE]}.\\  
A good answer should follow these rules:\\  
1. It should be in \texttt{[LANGUAGE]}, except when the instruction explicitly requests the answer in a different language.\\  
2. It should answer the request in the instruction.\\
3. It should be factually and semantically comprehensible.\\  
4. It should be grammatically correct and fluent.\\  
\\
\texttt{<instruction>}\\  
\texttt{[INSTRUCTION]} \\ 
\texttt{</instruction>} \\ 
\\
\texttt{<answer1>}  \\
\texttt{[OUTPUT1]} \\ 
\texttt{</answer1>}\\  
\\
\texttt{<answer2>}  \\
\texttt{[OUTPUT2]}\\  
\texttt{</answer2>} \\ 
\\
FIRST, provide a one-sentence comparison of the two answers, explaining which you prefer and why.\\  
SECOND, on a new line, state only `answer1' or `answer2' to indicate your choice. If both answers are equally good or bad, state `tie'. Your response should use the format:\\  
\\
Comparison: <one-sentence comparison and explanation> \\ 
\\
Preferred: <`answer1' or `answer2' or `tie'>  
\end{tcolorbox}

\end{document}